\documentclass[journal]{IEEEtran}
\usepackage{cite}

\ifCLASSINFOpdf
\else
\fi
\usepackage{amsmath}
\usepackage{algorithmic}
\usepackage{array}
\usepackage{stfloats}
\usepackage{url}
\usepackage{graphicx}
\usepackage{amsmath}
\usepackage{multirow}
\usepackage{booktabs}
\usepackage{makecell}
\usepackage{orcidlink}
\usepackage{textcomp}
\usepackage{soul}
\usepackage{xcolor}
\usepackage{ulem}
\usepackage{longtable}
\usepackage{threeparttable}
\usepackage{hyperref}
\begin{document}

\title{A Tale of Single-channel Electroencephalogram: Devices, Datasets, Signal Processing, Applications, and Future Directions}

\author{Yueyang Li \textsuperscript{\large\orcidlink{0009-0008-5310-124X}},
        Weiming Zeng\textsuperscript{\large\orcidlink{0000-0002-9035-8078}},~\IEEEmembership{Senior Member,~IEEE} ,
        Wenhao Dong\textsuperscript{\large\orcidlink{0009-0006-3764-7969}},
        Di Han\textsuperscript{\large\orcidlink{0009-0009-1869-5131}},
        Lei Chen\textsuperscript{\large\orcidlink{0009-0001-8562-6138}},
        Hongyu Chen\textsuperscript{\large\orcidlink{0009-0004-2934-3135}},
        Zijian Kang\textsuperscript{\large\orcidlink{0009-0005-5260-8805}},
        Shengyu Gong\textsuperscript{\large\orcidlink{0009-0001-9004-540X}},
        Hongjie Yan\textsuperscript{\large\orcidlink{0009-0000-2553-2183}}, 
        Wai Ting Siok\textsuperscript{\large\orcidlink{0000-0002-2154-5996}},
        and Nizhuan Wang\textsuperscript{\large\orcidlink{0000-0002-9701-2918}}

\thanks{This work was supported by the National Natural Science Foundation of
China (grant number 31870979), the Hong Kong Polytechnic University Start-up Fund (Project ID: P0053210) and the Hong Kong Polytechnic University Faculty Reserve Fund (Project ID: P0053738). (Corresponding author: Weiming Zeng, and
Nizhuan Wang)}
\thanks{Yueyang Li, Weiming Zeng, Wenhao Dong, Di Han, Lei Chen, Zijian Kang, Shengyu Gong and Hongyu Chen are with the Laboratory of Digital Image and Intelligent Computation, Shanghai Maritime University, Shanghai 201306, China (email: lyy20010615@163.com, zengwm86@163.com, jsxzdwh@163.com, handi@stu.shmtu.edu.cn, chen\_sanshi@163.com, hongychen676@gmail.com, 202430310078@stu.shmtu.edu.cn, gong2002417@126.com.)}
\thanks{Hongjie Yan is with the Department of Neurology, Affiliated Lianyungang Hospital of Xuzhou Medical University, Lianyungang 222002, China (email: yanhjns@gmail.com).}
\thanks{Wai Ting Siok and Nizhuan Wang are with the Department of Chinese and Bilingual Studies, The Hong Kong Polytechnic University, Hong Kong, SAR, China (e-mail: wai-ting.siok@polyu.edu.hk, wangnizhuan1120@gmail.com).}}

%
%

\markboth{Accepted by IEEE TRANSACTIONS ON INSTRUMENTATION AND MEASUREMENT}%
{Shell \MakeLowercase{\textit{et al.}}: Bare Demo of IEEEtran.cls for IEEE Journals}
%



\maketitle

\begin{abstract}
Single-channel electroencephalogram (EEG) is a cost-effective, comfortable, and non-invasive method for monitoring brain activity, widely adopted by researchers, consumers, and clinicians. The increasing number and proportion of articles on single-channel EEG underscore its growing potential. This paper provides a comprehensive review of single-channel EEG, focusing on development trends, devices, datasets, signal processing methods, recent applications, and future directions. Definitions of bipolar and unipolar configurations in single-channel EEG are clarified to guide future advancements. Applications mainly span sleep staging, emotion recognition, neurofeedback, educational research, and clinical diagnosis. Additionally, we discuss about the artificial intelligence (AI)-based EEG generation techniques, advancements through the integration of advanced signal processing with AI, innovations in hardware development, and strategies for the integration of wearables enabled by the Internet of Things (IoT), collectively establishing a foundational roadmap for future developments in single-channel EEG systems and their applications.

\end{abstract}


%
\IEEEpeerreviewmaketitle

\section{INTRODUCTION}
\begin{figure}
    \centering
    \includegraphics[width = 2.5in]{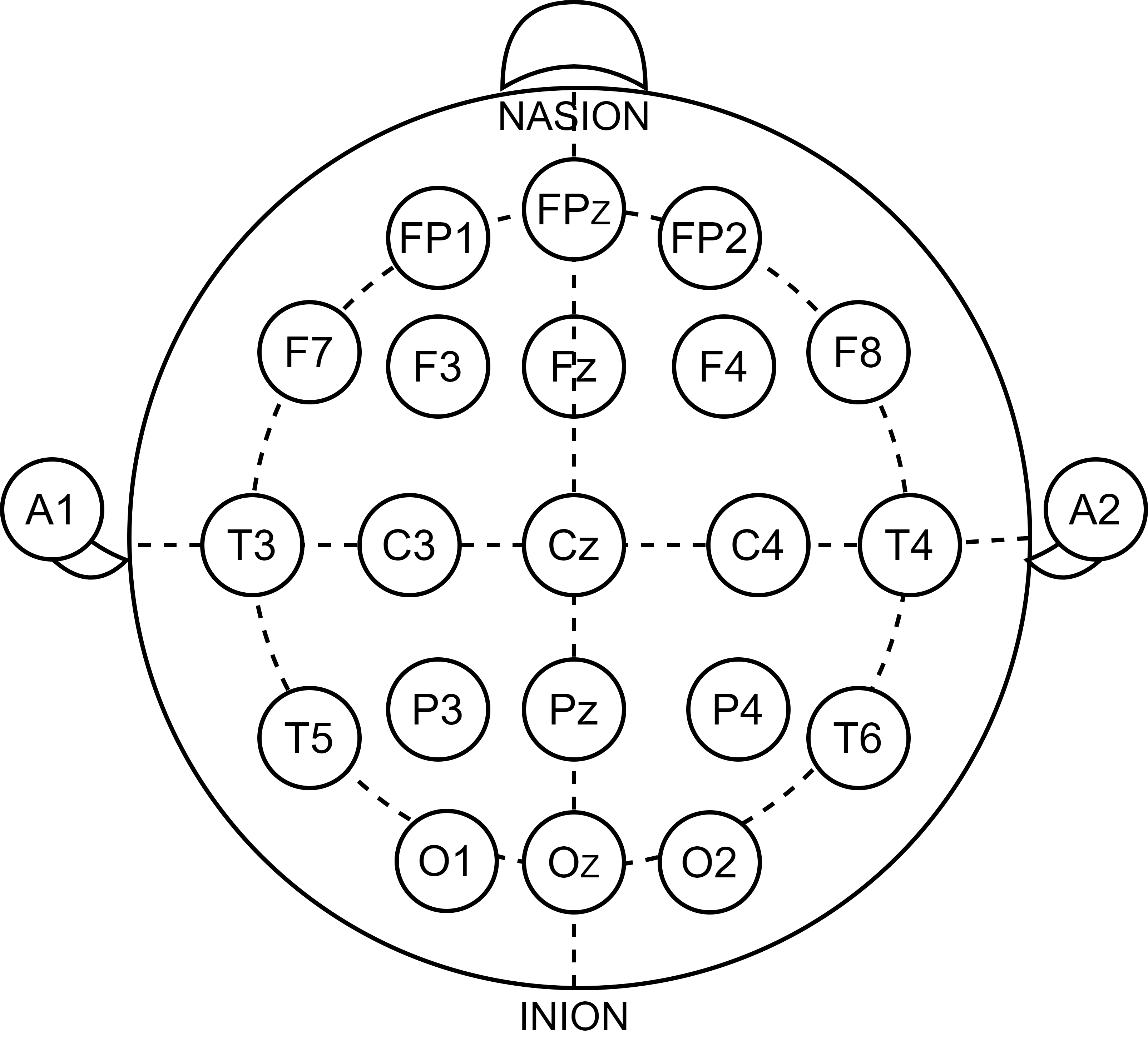}
    \caption{Scalp electrode placement in 10–20 system for EEG acquisition \cite{10-20}.}
    \label{10-20}
\end{figure}

%
%
%
%
\IEEEPARstart{E}{lectroencephalography} (EEG) is a technique for recording the brain's electrical signals, which has high temporal resolution and non-invasiveness, making it an indispensable tool in brain research and neurological disorder diagnosis \cite{EEG}. EEG measures the electrical activity by detecting voltage fluctuations resulting from ionic current flows within the neurons of the brain \cite{EEG-principle}. By recording electrical activity from the scalp, EEG can provide the real-time insights into brain function, enabling the investigation of cognitive processes under different conditions and mechanisms of neurological disorders \cite{EEG—attribute}. Traditionally, EEG signals are taken on the surface of the scalp, while the intracranial EEG (iEEG) directly records electrical signals inside the brain using implanted electrodes \cite{iEEG}. In this paper, we focus primarily on traditional scalp EEG signals, in which the International 10-20 system, shown in Fig. \ref{10-20}, is commonly used for EEG electrode placement and for correlating the external skull locations to the underlying cortical areas \cite{10-20}. 

In clinical settings, EEG is essential for diagnosing and monitoring conditions such as epilepsy and sleep disorders, as well as assessing brain function during cognitive tasks, aiding in the understanding of neural mechanisms underlying perception, memory, and decision-making \cite{epilepsy1, sleep-review1, depression1, skroke1,Parkinson1}. Moreover, EEG applications extend to non-clinical settings through its portability and cost-effectiveness, enabling emotion and stress monitoring for mental health management, attention tracking in educational environments, and concentration optimization in sports and meditation training \cite{emotioneeg,stress1,education-review1}. However, traditional EEG systems often employ multiple channels to capture a detailed spatial map of brain activity. High-density EEG systems enhance spatial resolution while introducing challenges in usability, convenience, and cost efficiency \cite{high-density-eeg}. Multi-channel EEG systems necessitate extensive skin preparation and adhesive conductive gels, resulting in time-consuming processes and user discomfort that impede repeated measurements \cite{multi-channel, wet}. Conductive gel application introduces signal quality variability and setup complexities in non-clinical environments \cite{skin}, while electrode caps induce physical discomfort, constraining recording duration and applicability for children and elderly subjects \cite{discomfort}. These limitations, combined with setup and maintenance complexities, pose significant barriers to widespread EEG adoption beyond laboratory settings \cite{MobileEEG}. In contrast, single-channel EEG, particularly with dry electrodes \cite{dry,dry2}, reduces setup complexity and enhances user comfort, thus making EEG more accessible for broader applications.

\begin{figure}
    \centering
    \includegraphics[width = 2.6in]{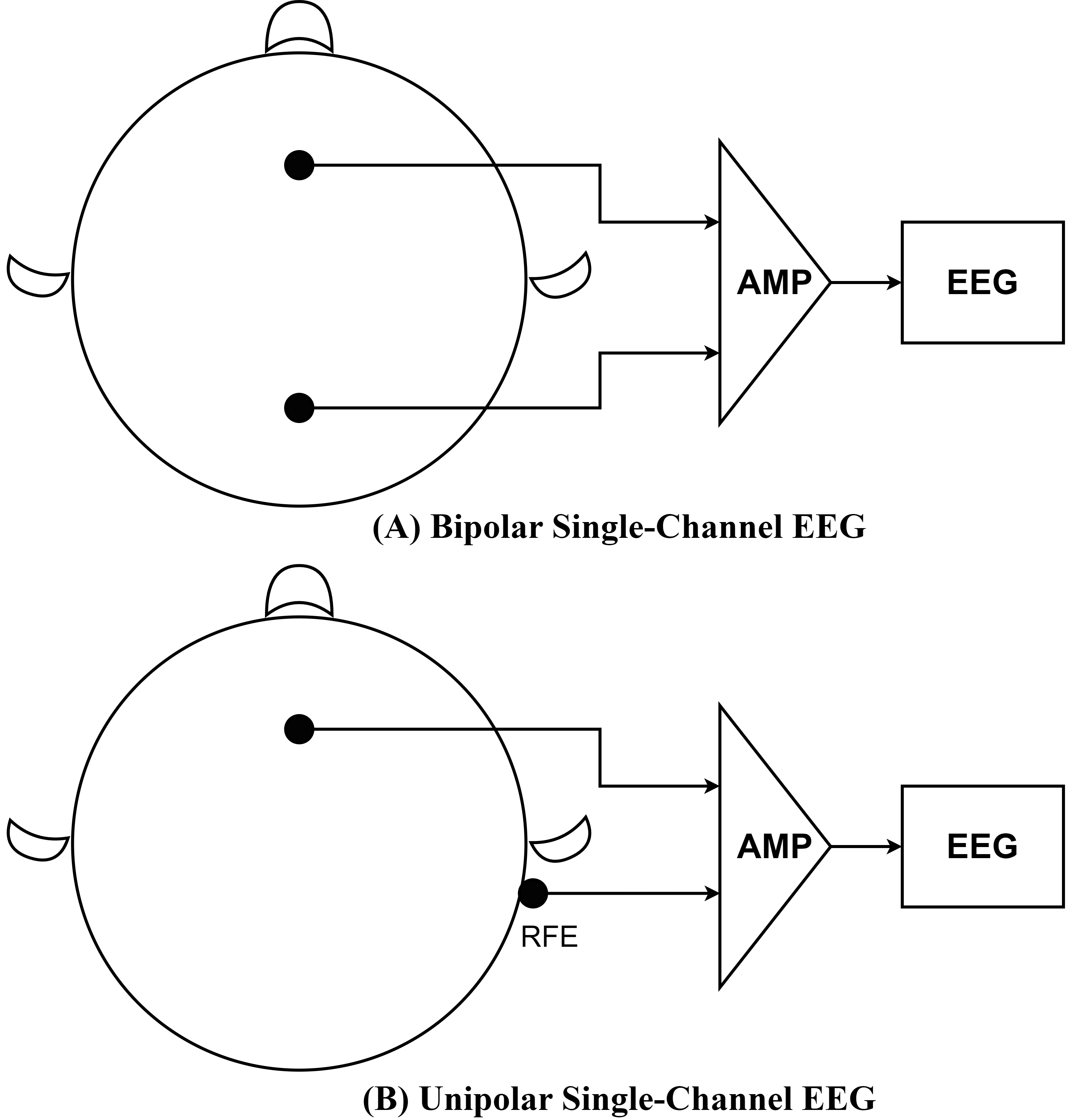}
    \caption{Illustration of bipolar/unipolar single-channel EEG \cite{polar}.}
    \label{polar}
\end{figure}

Single-channel EEG is a simple system that applies two electrodes to record single-channel electrical activity, which can be classified into unipolar single-channel EEG and bipolar single-channel EEG based on the electrode placement and reference selection \cite{single-channel-def}, as shown Fig. \ref{polar}. Bipolar single-channel EEG means that two electrodes are placed at different locations on the scalp and the potential difference between these two electrodes reflects the difference in electrical activity between two specific brain areas. The unipolar reference is used when all the electrodes are referenced to a unique physical reference or a unique virtual reference \cite{UnipolarReferences}. The physical reference is typically an electrode (e.g., Cz, Fz and Oz) placed on the scalp or body surface during the recording setup. The virtual reference is a linear combination of recordings from all electrodes, usually obtained during offline processing after EEG data acquisition. Typical examples of virtual references include the reference electrode standardization technique (REST), average reference (AR), and linked-mastoids/ears reference (LM) \cite{REST,RV,LM}. The primary difference is that unipolar single-channel EEG employs one recording electrode and one reference electrode, whereas bipolar single-channel EEG uses two recording electrodes to measure the relative potential changes between them. Single-channel EEG systems comprise wired variants, predominantly integrated within multi-channel systems, and wireless variants, primarily implemented in portable devices. Single-channel EEG enables critical health monitoring applications, particularly in sleep stage analysis and sleep apnea detection, while facilitating non-intrusive home monitoring through wearable device integration \cite{SHNN,SingleChannelNet,Single-channelEEGsleepstage}. Beyond sleep studies, this technology detects brainwave pattern changes associated with stress and relaxation states for stress reduction interventions, and measures brain activity linked to mental effort for cognitive assessment in educational settings and user experience research \cite{stress2,cognitive1}.

As a simplified EEG recording technique, single-channel EEG shows significant potential in clinical and non-clinical applications \cite{biondi2022noninvasive}. Balam et al.\cite{drowsiness_review} systematically reviewed single-channel EEG-based drowsiness detection methods, addressing safety implications and analytical techniques while excluding broader single-channel EEG applications. An early mini-review \cite{lin2010review} explored wireless and wearable EEG systems and brain-computer interfaces, primarily focusing on broader technological advancements rather than single-channel applications. A specialized study examining ocular artifact denoising in single-channel EEG signals advanced signal quality enhancement \cite{artifacts_review}, yet maintained a narrow focus excluding broader applications and configurations. Despite the comprehensive analysis of multi-channel systems \cite{anders2022wearable}, the unique characteristics of single-channel EEG system configurations remain relatively unexplored. Therefore, a narrative review is essential for summarizing research on single-channel EEG, highlighting its devices, datasets, applications, challenges, and future development directions, thereby addressing the growing importance of this technology.

A literature search was conducted in the Google Scholar database up to October 2024 using the search terms {``}Single-channel" AND {``}EEG" OR {``}electroencephalography" OR {``}devices"; {``}Single-channel" AND {``}EEG" OR {``}electroencephalography" OR {``}datasets"; {``}Single-channel" AND {``}EEG" OR {``}electroencephalography" OR {``}signal processing"; {``}Single-channel" AND {``}EEG" OR {``}electroencephalography" OR {``}applications". Additionally, we searched for {``}single-channel EEG" AND {``}sleep staging" OR {``}emotion recognition" OR {``}education research" OR {``}clinical diagnosis" OR {``}EEG signal generation" OR {``}bipolar" OR {``}unipolar" OR {{``}neurofeedback" OR {``}AI" OR {``}deep learning" OR {``}generative model" OR {``}hardware" OR {``}IoT"}. We also retrieved relevant articles cited in other narrative reviews and screened references from original papers to ensure no significant reports were overlooked. We included articles that specifically discussed single-channel EEG in terms of its technical development, usage in different contexts, advancements in signal processing, and future potential.

In Section II, we review the evolution of EEG and explore research trends in single-channel EEG. Section III addresses consumer-grade single-channel EEG device design and implementation. Section IV presents datasets utilized in single-channel EEG research. Section V analyzes signal processing methodologies. Section VI investigates single-channel EEG applications encompassing sleep staging, disease diagnosis, emotion recognition, neurofeedback, and educational research. Section VII evaluates the preceding content while examining challenges and future development trajectories in single-channel EEG research. Section VIII concludes by addressing implications for single-channel EEG technology advancement.

\section{TRENDS OF SINGLE-CHANNEL EEG}
Single-channel EEG systems are advancing significantly across various fields due to their practicality, affordability, and versatility. Enhanced by wearable technology, these systems are widely used in personal health monitoring, meditation aids, and sleep tracking, providing insights into brain activity patterns and biofeedback for therapeutic purposes \cite{trend1}. They are also crucial in cognitive and educational research, offering simplified yet effective tools favored in student projects and educational settings. Technological improvements in signal processing have boosted their accuracy and reliability, enhancing their efficacy across diverse applications \cite{wanghsu,lin}. Integration with other wearable sensors and mobile applications expands their functionality, offering comprehensive physiological and psychological insights. Commercially, devices like the NeuroSky MindWave and Sichiray demonstrate a growing market presence, with improved user interfaces and comfort for extended wear \cite{trend3}.

\begin{figure}
    \centering
    \includegraphics[width = 3.2in]{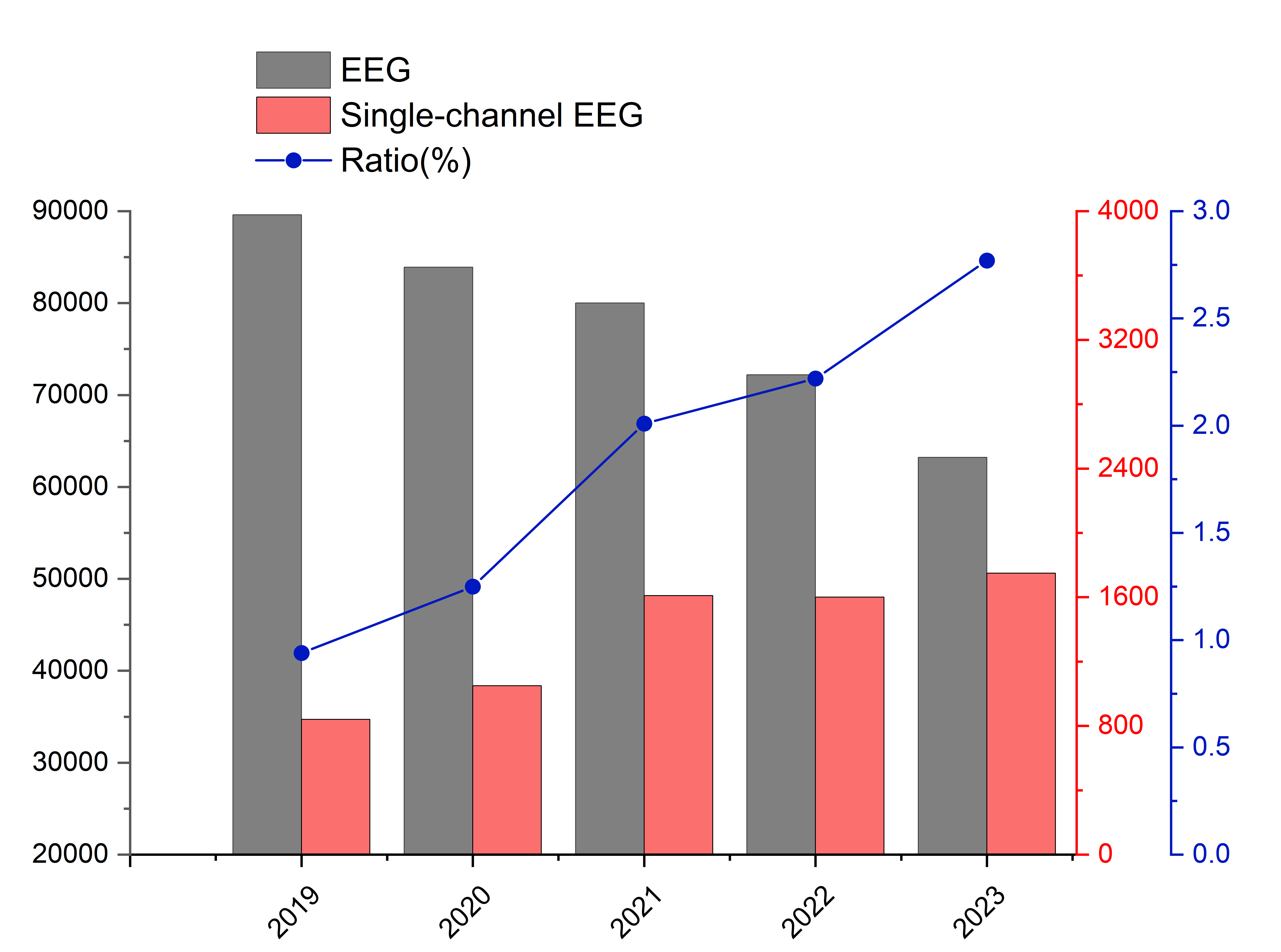}
    \caption{Total number of general and single-channel EEG publications over the past 5 years, as searched on Google Scholar. Single-channel EEG is an increasingly popular research topic.}
    \label{trend}
\end{figure}

Although our literature search encompassed publications up to October 2024, the incomplete volume of articles for 2024 was taken into account by focusing our trend analysis on studies from 2019 to 2023. The total number of articles containing the keyword {``}Single-channel EEG" from 2019 to 2023 was retrieved from Google Scholar. All studies related to EEG were identified using the keyword {``}EEG". Non-relevant articles, where the search keyword appeared only as part of the author's name, were excluded. The survey revealed an increasing focus on single-channel EEG studies between 2019 and 2023, as depicted in Fig. \ref{trend}. The number of publications on single-channel EEG rose from 841 in 2019 to 1,050 in 2021, reaching 1,750 in 2023. The ratio of Single-channel EEG to general EEG studies increased from 0.94\% in 2019 to 2.01\% (2021) and 2.77\% (2023). This represents a nearly six-fold increase in the proportion of single-channel EEG studies over the period, reflecting growing recognition and attention from researchers. Furthermore, advancements in signal processing and data analysis methods for single-channel EEG have significantly contributed to the rapid development of this field \cite{signalpro}.

\section{DEVICES of SINGLE-CHANNEL EEG}
This section introduces the commonly used single-channel EEG devices in detail, focusing on their instrumentation and measurement characteristics, including features such as sensor, analog-to-digital converter (ADC) resolution, sampling rate, connectivity, battery life, {max power consumption} and data availability.
\renewcommand\arraystretch{1.5}
\begin{table*}[!ht]
    \centering
    \caption{DEVICES of SINGLE-CHANNEL EEG}
\begin{tabular}{p{2.0cm}<{\centering}p{0.8cm}<{\centering}p{0.8cm}<{\centering}p{1.3cm}<{\centering}p{1.1cm}<{\centering}p{1.3cm}<{\centering}p{1.4cm}<{\centering}p{0.7cm}<{\centering}p{1.15cm}<{\centering}p{1.65cm}<{\centering}p{1.3cm}<{\centering}}
    
    \Xhline{1.2pt}
    \textbf{Device} & \textbf{Released} & \textbf{Sensor} & \textbf{ADC Resolution (bits)} & \textbf{Sampling Rate (Hz)} & \textbf{Electrode Type} & \textbf{Connection} & \textbf{Price (USD)} & \textbf{Battery Life (Hours)} & \textbf{{Max Power Consumption}} & \textbf{Data Available?} \\
    
    \Xhline{1.2pt}
    \textbf{NeuroSky MindSet} \cite{NeuroSkyMindSet} & 2007 & Fp1 & 12 & 512 & Dry & Bluetooth & 199.99 & {-} & {15mA @ 3.3V} & Y \\ \hline
    \textbf{NeuroSky Mindflex} \cite{NeuroSkyMindflex} & 2009 & Fp1 & 12 & 512 & Dry & Wired & 79.99 & {-} & {15mA @ 3.3V} & N \\ \hline
    \textbf{NeuroSky MindWave} \cite{MindWave} & 2011 & Fp1 & 12 & 512 & Dry & Bluetooth & 99.99 & {-} & {15mA @ 3.3V} & Y\\ \hline
    \textbf{XWave headset} \cite{XWave} & 2011 & Fp1 & 12 & 512 & Dry & Wired & 99 & {-} & {15mA @ 3.3V} & N\\ \hline
    \textbf{NeuroSky MindWave Mobile} \cite{MindwaveMobile} & 2012 & Fp1 & 12 & 512 & Dry & Bluetooth & 129.99 & {-} & {15mA @ 3.3V} & Y\\ \hline
    \textbf{NecoMimi} \cite{MyndPlay} & 2012 & Fp1 & 12 & 512 & Dry & Bluetooth & 69 & {-} & {15mA @ 3.3V} & N\\ \hline
    \textbf{Olimex EEG-SMT} \cite{Acampora} & 2013 & Fpz or Oz & 10 & 256 & Dry & - & - & 3 & {-} & Y\\ \hline
    \textbf{MyndPlay Mindband} \cite{MyndPlay} & 2014 & Fp1 & 12 & 512 & Dry & BLE & 219 & 10 & {15mA @ 3.3V} & Y\\ \hline
    \textbf{Aurora Dreamband} \cite{Aurora} & 2015 & Fp1 & - & - & Dry & BLE & 299 & 8 & {15mA @ 3.3V} & Y\\ \hline
    \textbf{NeuroSky MindWave Mobile II} \cite{MindwaveMobile2} & 2018 & Fp1 & 12 & 512 & Dry & Bluetooth/BLE & 129.99 & {-} & {15mA @ 3.3V} & Y\\ \hline
    \textbf{Macrotellect Brainlink lite V2} \cite{Pro} & - & Fp1 & 12 & 512 & Dry & Bluetooth & 179 & 5 & {15mA @ 3.3V} & Y\\ \hline
    \textbf{Macrotellect Brainlink Pro} \cite{Pro} & - & Fp1 & 12 & 512 & Dry & BLE & 259 & 4 & {15mA @ 3.3V} & Y\\ \hline
    \textbf{Sichiray 2.0} \cite{Sichiray} & 2022 & Fp1 & 12 & 215 & Dry/Wet & BLE & 80 & 8 & {5mA @ 3.3V} & N\\
    \Xhline{1.2pt}
\end{tabular}
\label{device}

\begin{tablenotes}
    \item[] \textbf{BLE}: Bluetooth Low Energy, a power-efficient version of Bluetooth. \textbf{AAA}: Standard AAA batteries are used to power the device. \textbf{Y}: Data is available for access or download. \textbf{N}: Raw data is unavailable for access or download. \textbf{-}: No relevant information found.
\end{tablenotes}
\end{table*}

EEG devices utilize two main types of electrodes: wet and dry \cite{dry}. Wet electrodes require conductive gel to reduce impedance at the skin-electrode interface, ensuring high-quality signal acquisition and are considered the gold standard \cite{wet}. However, the need for gel increases preparation time and limits device portability. In contrast, dry electrodes are made from materials such as conductive polymers or metals that can capture electrical signals from the scalp without the need for conductive gel \cite{dry2}. This design makes dry electrodes more convenient and user-friendly, particularly for single-channel EEG devices. {Additionally, semi-dry electrodes represent a compromise between wet and dry types, using minimal electrolyte to improve signal quality while maintaining relative convenience.} Among these options, dry electrodes are widely used as they simplify setup, enhance portability and suit wearable and long-term monitoring.

NeuroSky was one of the pioneering companies to produce and distribute consumer-grade EEG devices in the market. In 2007, NeuroSky launched their first product, the MindSet headset, which utilizes a dual-electrode system with an ear-mounted reference and forehead-positioned signal capture electrode \cite{NeuroSkyMindSet}. {Building on this initial success, NeuroSky expanded their product line with more advanced devices like the MindWave}, a cost-effective single-channel EEG solution that incorporates the ThinkGear ASIC Module (TGAM) \cite{tgam}, {a specialized chip featuring 12-bit ADC resolution, 512Hz sampling rate and 15mA @ 3.3V power consumption}, to measure and output EEG power spectra such as $\alpha$, $\beta$ waves and derived metrics (attention, meditation), providing an easy-to-use brainwave monitoring solution. Additionally, The NeuroSky MindWave consists of a T-shaped headband with wide ear clips and forehead sensors above the eyes, connects via Bluetooth, and operates for 8 hours using a single AAA battery \cite{MindWave}. The NeuroSky MindWave Mobile improves upon the original with better comfort and connectivity, featuring automatic wireless pairing, enhanced ergonomics, and compatibility with various educational and gaming apps from their App Store \cite{MindwaveMobile}. {The MindWave Mobile II further enhances comfort through its TGAM2.9B Bluetooth v4.0 module, flexible rubber sensor arm, rounded forehead sensor tip, T-shaped headband and wide ear clips, while offering 8-hour battery life from a single AAA battery and compatibility with Windows, Mac, iOS and Android platforms \cite{MindwaveMobile2}. For developers interested in custom implementations}, NeuroSky provides the TGAM module at \$49 per piece, offering comprehensive measurement capabilities including blink detection, mental effort, familiarity, appreciation, emotional spectrum, creativity, and alertness measurement \cite{huangliu}. {This core technology has been widely implemented, powering both NeuroSky's own products (MindSet \cite{NeuroSkyMindSet}, MindWave series \cite{MindwaveMobile}, Mindflex \cite{NeuroSkyMindflex}) and third-party devices (XWave headset \cite{XWave}, NecoMimi \cite{MyndPlay}, MyndPlay Mindband \cite{MyndPlay}, Macrotellect Brainlink \cite{Pro}). While based on the same TGAM architecture, these devices are optimized for distinct application domains, ranging from cognitive research and educational assessment to interactive entertainment systems.} For instance, Mindflex \cite{NeuroSkyMindflex} enables users to control a foam ball through obstacles using brainwaves, while NecoMimi \cite{MyndPlay} features plush cat ears that respond to brainwave activity. In a similar market segment, the Sichiray 2.0 provides consumer and educational EEG monitoring with a compact design, operating at 3.3V @ 5mA via Bluetooth communication. The device incorporates a micro USB charging port, 3-125 Hz frequency range, 12-bit ADC, and 215 Hz sampling rate, outputting eSense metrics including $\delta$ and $\theta$ waves for mental state analysis while utilizing an onboard notch filter for 50 Hz interference elimination \cite{Sichiray}.

Technological advancements have driven single-channel device proliferation, leading manufacturers to develop diverse consumer-oriented solutions. MyndPlay's TGAM-based headband integrates recreational features with cognitive development functionalities \cite{MyndPlay}, while the Aurora Dreamband \cite{Aurora} incorporates sensor technology for movement and sleep cycle monitoring. These consumer-grade devices facilitate cognitive state research without extensive technical expertise \cite{dry2} while expanding into education and gaming applications \cite{consumer}. Their economic accessibility enhances user adoption while maintaining core functionalities. Despite performance constraints in attention and meditation studies, environmental calibration integration, multi-sensor fusion, and deep learning technologies have enhanced their educational adaptability. NeuroSky devices demonstrate educational research efficacy despite simplified signal acquisition \cite{education-review1}. Compared to multi-channel systems such as Emotiv \cite{Emotiv}, Muse \cite{muse}, and OpenBCI \cite{OpenBCI}, single-channel devices offer enhanced convenience for entry-level research requiring moderate precision. The BrainLink Lite V2 and Pro \cite{Pro} implementations utilize single-channel (Fp1) capabilities for raw EEG acquisition and eSense-based monitoring. The Pro variant features specialized earclips, enhanced relaxation protocols, cardiac monitoring, EmoLight functionalities, and BLE 4.0 connectivity, surpassing Lite's Bluetooth 3.0. Pro additionally incorporates PPG heart rate sensors and expanded BCI capabilities, offering comprehensive accessory packages versus Lite's basic configuration. Additionally, the Pro model integrates PPG heart rate sensors and expanded BCI capabilities, with both variants offering mobile app compatibility, though Pro provides a more comprehensive accessory package versus Lite's basic headband configuration. Additionally, The Olimex EEG-SMT utilizes a 256 Hz sampling rate and two-channel differential input 10-bit ADC, though operationally employing a single channel configuration with dual active electrodes (CH- and CH+) for voltage differential measurement \cite{Acampora}.

In the field of instrumentation and measurement, single-channel devices will maintain significance as algorithms and sensor technology advance, with future developments focusing on intelligent signal processing optimization and enhanced device adaptability and accuracy. User experience and usability factors, including wearing comfort for long-term use, wireless connectivity for simplified data transmission, and compact design for improved portability, are critical considerations that influence their widespread application. {Table \ref{device} summarizes the key characteristics of single-channel EEG devices documented in academic literature, covering specifications from released time and sensor type to ADC resolution, sampling rate, electrode configuration, connectivity options, pricing, battery duration, and data accessibility. These devices exhibit varying technical capabilities, particularly in their ADC resolution and sampling rates, while their choice of connection method (Bluetooth or wired) further impacts the quality of signal capture and efficiency of data transmission - factors whose importance varies depending on specific research needs.} Looking forward, comprehensive user studies focusing on comfort, operational simplicity, and overall satisfaction will be essential to guide future design optimizations and enhance the user experience.

\section{DATASETS of SINGLE-CHANNEL EEG}
As stated in Section I, this review refers to two types of single-channel EEG: unipolar and bipolar EEG, depending on the reference electrode position. Bipolar single-channel EEG signals typically exist within multi-channel EEG data and are rarely processed or labeled separately. Researchers select specific electrode pairs that form the channel based on prior knowledge and the requirements of downstream tasks. This personalized and task-specific selection hinders the establishment and sharing of standardized single-channel EEG datasets. In addition, the experimental design of multi-channel EEG datasets often differs from that of single-channel applications. For instance, most bipolar single-channel EEG data used in sleep staging are derived from specific bipolar channels within the polysomnography (PSG) data \cite{Alightweight}. Given the diverse and unstandardized nature of the data generated by bipolar single-channel EEG, this article focuses solely on unipolar single-channel EEG datasets, which are scarce in public repositories. Table \ref{data} provides a summary of the details of the public single-channel EEG datasets.

The K-EmoCon dataset \cite{K-EmoCon} is a multimodal sensor dataset designed for continuous emotion recognition in natural conversations. It addresses the limitations of existing emotion datasets in capturing individual emotions in natural environments. K-EmoCon includes audio and video recordings, EEG and peripheral physiological signals collected from 32 participants during 16 debates, each approximately 10 minutes long. The EEG signals were collected using the NeuroSky MindWave headset, which uses two dry sensor electrodes: one on the forehead (FP1 - 10/20 system) and the other on the left earlobe (reference). The data collection process consists of four stages: entry, baseline measurement, debate and emotion annotation. While watching the debate video, participants annotated their emotions every 5 seconds, including self-annotation, opponent annotation and external observer annotation. The annotations cover arousal-valence and 18 emotion categories. The data has been thoroughly checked and preprocessed to ensure signal integrity and device reliability. 

The Emo\_Food dataset \cite{Emo-Food} was created from an EEG experiment involving 20 random volunteers to study the effects of food images with different emotional valences on EEG activity. Participants viewed positive, neutral, and negative food images, with a 10-second relaxation period between each trial. The EEG data were recorded at a sampling rate of 512 Hz using a NeuroSky MindWave Mobile II, with the sensor placed on the left forehead. This dataset offers valuable data for research in affective computing, neuroscience and psychology.

Acampora et al. proposed a single-channel EEG dataset from a portable steady-state visual evoked potential (SSVEP) based BCI involving 11 volunteers \cite{Acampora}. The recorded EEG data from a single volunteer contains the response to an intermittent source of light, which is emitted at four different frequencies, namely 8.57 Hz (F1), 10 Hz (F2), 12 Hz (F3) and 15 Hz (F4). At the end of the experiment, a total of 44 recordings of 16 s each one sampled at 256 Hz, for a total of 180224 samples was obtained. This dataset serves as a crucial baseline for single-channel EEG, particularly in the emerging field of human-robot interaction.

The scarcity of single-channel EEG datasets poses a significant challenge in current neuroscience research, and expanding these datasets is essential given their advantages over multi-channel EEG datasets in equipment cost, data processing complexity and practical application \cite{cost1}. However, expanding these datasets requires addressing challenges in data collection, annotation and data quality control.

\begin{table*}[!htbp]
	\centering
	\caption{Single-channel EEG Datasets with Additional Details}
	\begin{tabular}{p{2.5cm}<{\centering} p{3cm}<{\centering} p{2.8cm}<{\centering} p{1.5cm}<{\centering} p{1.5cm}<{\centering} p{3.5cm}<{\centering}}
  \Xhline{1.2pt}
\textbf{Dataset} & \textbf{EEG Device} & \textbf{Collected Data} & \textbf{Sampling Rate} & \textbf{Subjects} & \textbf{Signal Duration} \\ \hline
\multirow{2}{*}{\textbf{K-EmoCon \cite{K-EmoCon}}} & \multirow{2}{*}{\begin{tabular}[c]{@{}c@{}}NeuroSky MindWave\\ Headset\end{tabular}} & Brainwave (Fp1 Channel EEG) & 125 Hz & \multirow{2}{*}{32} & \multirow{2}{*}{\begin{tabular}[c]{@{}c@{}}Approx. 10 min \\ per session\end{tabular}} \\ \cline{3-4}
&  & Attention \& Meditation & 1 Hz &  &  \\ \hline
\textbf{Emo\_Food \cite{Emo-Food}} & \begin{tabular}[c]{@{}c@{}}NeuroSky MindWave\\ Mobile II\end{tabular} & Brainwave (Fp1 Channel EEG) & 512 Hz & 20 & \begin{tabular}[c]{@{}c@{}}3 trials per subject\\ (duration not specified)\end{tabular} \\ \hline
\textbf{SSVEP\_BCI \cite{Acampora}} & Olimex EEG-SMT & Brainwave (Fpz or Oz Channel EEG) & 256 Hz & 11 & Approx. 64 s per subject \\
\Xhline{1.2pt}
	\end{tabular}
	\label{data}
\end{table*}



\section{SIGNAL PROCESSING of SINGLE-CHANNEL EEG}
\begin{figure*}
    \centering
    \includegraphics[width = 1\textwidth]{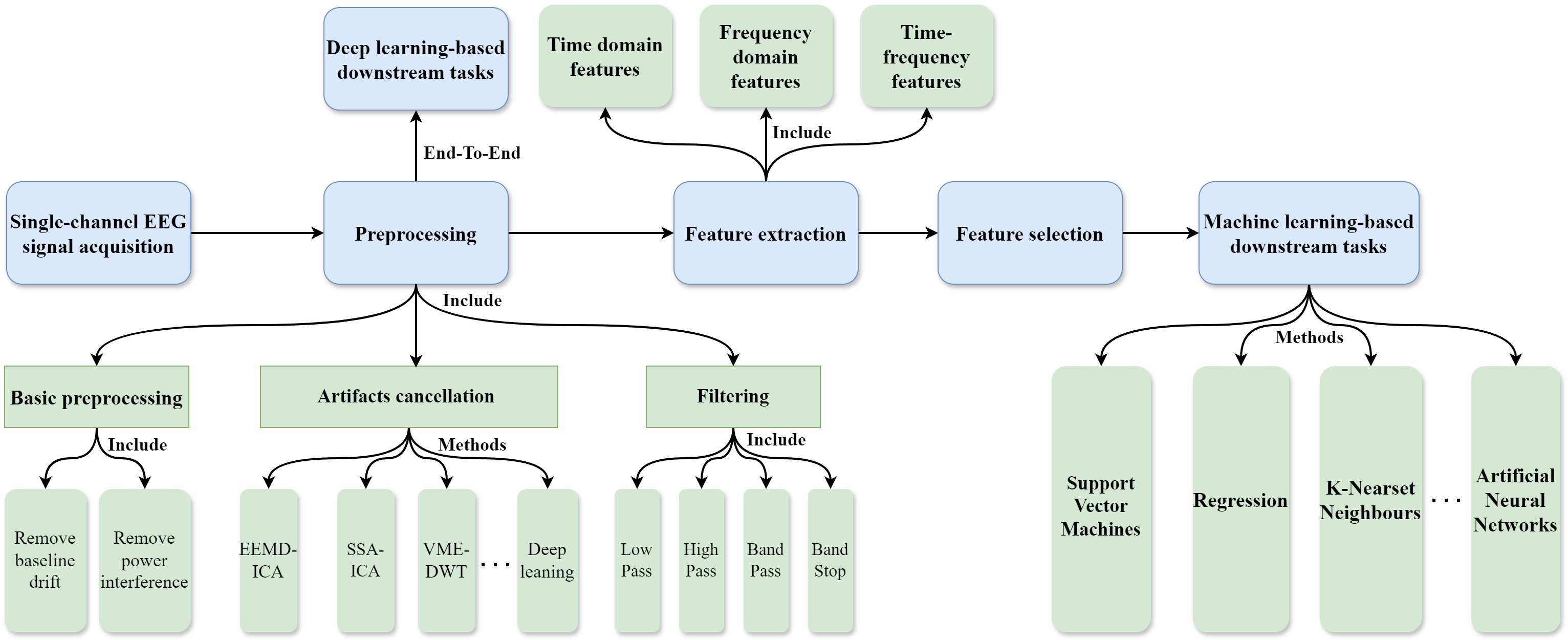}
    \caption{Flowchart of single-channel EEG signal processing illustrating key steps from signal acquisition to downstream tasks, encompassing preprocessing, feature extraction, feature selection, as well as traditional machine learning and deep learning methods.}
    \label{signalprocess}
\end{figure*}

In this section, we present a detailed processing framework of single-channel EEG signals, encompassing acquisition, preprocessing, feature extraction, feature selection and ML or DL-based downstream tasks, as depicted in Fig. \ref{signalprocess}. 
\subsection{Signal Preprocessing}
\subsubsection{Basic Preprocessing \& Filtering}
In the preprocessing stage of single-channel EEG signals, removing baseline drift and power supply interference is a key pre-processing step. Baseline drift in EEG signals refers to a type of low-frequency signal variation caused by poor electrode contact or other slowly changing factors. High-pass filtering or empirical mode decomposition (EMD) methods are usually used to remove low-frequency drifts, including baseline drifts \cite{drift}. To remove power supply interference, notch filters or adaptive filtering are often applied to specifically eliminate power frequency interference of 50 Hz or 60 Hz \cite{power}. Based on the designed filters, single-channel EEG signals are usually divided into five different frequency bands: $\delta$ (0.5-4 Hz), $\theta$ (4-7 Hz), $\alpha$ (8-13 Hz), $\beta$ (14-30 Hz), and $\gamma$ (\textgreater 30, but usually \textless 100 Hz) \cite{zhou2020automatic}. 

\subsubsection{Artifacts Cancellation}
Single-channel EEG artifact cancellation faces critical challenges, including the broad spectral impact of artifacts, the lack of standardized preprocessing steps that effectively filter noise while preserving signal integrity and high variability among subjects and trials in phase-locked EEG signals. These artifacts mainly include electrooculogram (EOG) artifacts, electromyography (EMG) artifacts, and motion artifacts \cite{artifact1}. EOG artifacts from blinking and eye movement appear as low-frequency, high-amplitude signals \cite{artifact2}, EMG artifacts from scalp and facial muscle activity show wide frequency range and high amplitude \cite{artifact3}, while motion artifacts due to poor electrode contact produce low-frequency, large-amplitude noise \cite{motion}. These artifacts can significantly reduce the accuracy of single-channel EEG data, increase the risk of misdiagnosis and complicate data processing. For instance, eye movement artifacts may be mistaken for epileptic seizures, while EMG artifacts may obscure genuine EEG abnormalities, leading to missed or incorrect diagnosis. These artifacts also limits the extensive application of single-channel EEG in portable and mobile devices, particularly in scenarios requiring high-precision. 

Since single-channel EEG systems lack multi-channel information commonly used in multi-channel EEG artifact removal methods such as independent component analysis (ICA), canonical correlation analysis (CCA) and independent vector analysis (IVA) \cite{artifact1}, they cannot be directly applied. These methods often fall under the category of blind source separation (BSS) \cite{bbs}, a technique used to separate a set of mixed signals into their original components without prior knowledge of the source signals or the mixing process. Some studies have applied decomposition methods such as ensemble empirical mode decomposition (EEMD) \cite{artifact5} and singular spectrum analysis (SSA) \cite{artifact6} to convert single-channel EEG signals into a multi-dimensional format. Hybrid methods have also been employed, where EEMD-ICA \cite{artifact7} decomposes signals into spectral independent modes, while SSA-ICA \cite{artifact8,artifact9} decomposes signals into multiple sub-signals, both followed by ICA to extract statistically independent sources for artifact removal. To avoid the edge effect in EEMD-ICA, Guo et al. proposed combining local mean decomposition (LMD) and ICA (LMD-ICA) \cite{artifact10}, although parameter adjustment in LMD is challenging. Inuso et al. proposed a method called wavelet-Independent Component Analysis (w-ICA) for EEG signal denoising that combines discrete wavelet transform {(DWT)} and ICA \cite{artifact12}. In many biomedical applications based on single-channel EEG, artifacts often appear in narrow frequency bands \cite{maddirala2021eye}. Thus, hybrid methods including ICA are more suitable for removing stereotyped artifacts, although ICA-based hybrid methods do not perform well due to the significant variations in multiple types of artifacts \cite{artifact13} and will shorten the detection time and reduce the battery life because of the high computational cost, which results in increased power consumption. In contrast, the low computational complexity of CCA extends the working time of single-channel EEG devices. A technique combining EEMD and CCA was proposed in \cite{artifact14,artifact18} to eliminate motion artifacts in single-channel EEG signals. This technique first uses EEMD to decompose the single-channel signal into a set of oscillatory components (intrinsic mode functions, IMFs), converting it into a multivariate signal. CCA then extracts the source signal from the multivariate data and its delayed version, with ICA-based methods typically used for EOG artifacts and CCA and IVA better suited for EMG artifacts, though no single method can completely remove all types of artifacts.

Compared with CCA, {wavelet transform (WT)} is a lower computational complexity method that can be used. WT represents the time-frequency pattern of a signal by decomposing it into multiple groups of coefficients. Compared to BSS, WT has a lower computational cost, making it more suitable for online applications of single-channel EEG. A method combining DWT and adaptive noise cancellation (ANC) was developed to remove eye movement artifacts from EEG \cite{artifact15}. Khatun et al. systematically evaluated the effectiveness of unsupervised WT decomposition techniques, specifically DWT and stationary wavelet transform (SWT), using four wavelet basis functions (haar, coif3, sym3, and bior4.4) and two thresholding methods (Universal Threshold and Statistical Threshold) for ocular artifact removal from single-channel EEG data \cite{artifact16}. Tengtrairat et al. proposed an online adaptive source separation method using pseudo-stereo noisy mixing and adaptive spectrum amplitude estimation \cite{artifact17}. Subsequent studies integrated various techniques: overlap segmentation with adaptive singular spectrum analysis (ASSA) and ANC \cite{artifact19}, Variational Mode Extraction-DWT (VME-DWT) algorithm for eye blink detection \cite{artifact20}, and combination of SSA, ICA, and SWT for EOG artifact removal \cite{artifact9}.

In addition, DL has been proposed to effectively capture artifacts in EEG signals. Lu et al. introduced Dual-Stream Attention-TCN for removing EMG artifacts from single-channel EEG signals \cite{artifact-dl-1}. This approach employs two parallel streams processing EEG signals in different frequency bands: the first stream focuses on extracting low-frequency EEG features to mitigate EMG interference, while the second stream integrates high-level features from the first stream with its own low-level features for comprehensive EEG reconstruction across the frequency spectrum. Cui et al. proposed EEGIFNet, a dual-branch interactive fusion neural network, for artifact removal in single-channel EEG signals \cite{artifact-dl-2}. EEGIFNet predicts clean EEG signals and residual artifacts through two branches, facilitating information exchange and fusion via interactive and fusion modules to effectively eliminate various types of artifacts. Zhang et al. presented a two-stage intelligent multi-type artifact removal model based on GRU autoencoder. This model aims to remove mixed multi-type artifacts from single-channel EEG by first identifying and then removing artifacts \cite{artifact-dl-3}. It employs attention mechanisms and adaptive feature extraction in the encoding feature domain to distinguish contaminated and clean EEG signals, achieving efficient artifact removal.

While traditional signal processing approaches like BSS and wavelet-based methods have established foundations for artifact removal, the field has evolved toward hybrid architectures and DL solutions that offer enhanced adaptability and robustness. Nevertheless, the optimal balance between computational efficiency and removal accuracy remains a crucial consideration for practical applications, especially in resource-constrained portable devices.

\subsection{Feature Extraction}
Time domain feature extraction focuses on deriving information directly from the EEG signals \cite{al2014methods}. These features include statistical properties (mean, variance, skewness, kurtosis) and signal characteristics (amplitude, duration, waveform dynamics), effectively capturing transient behaviors and dynamic traits within EEG signals. In addition to fundamental statistical features, more intricate aspects can be revealed through pattern recognition techniques applied to time series data, such as tracking changes in frequency components over time using methods like short-time Fourier transform (STFT) \cite{gu2021eeg,zhang2019spectral}.

Frequency domain analysis examines the distribution of energy within EEG signals across different frequency bands. Key features include power spectral density (PSD), band power, spectral entropy, and energy ratios among frequency bands. Fourier transform or WT enables the transformation of time domain signals into frequency domain representations, facilitating extraction of frequency domain features that reflect brain activity characteristics. In addition to traditional PSD analysis, advanced methods like nonlinear spectrum estimation (e.g., Hilbert-Huang transform) can be applied to unveil intrinsic signal patterns \cite{zhaotemporal,zhang2019spectral}. 

Time-frequency domain analysis integrates the strengths of both time and frequency domains, employing techniques like WT to examine signal characteristics across different time intervals and frequencies. This approach is well-suited for analyzing non-stationary signals such as single-channel EEG, offering insights into localized time-frequency characteristics of the signal \cite{hosse}.

\subsection{Feature Selection}
Feature selection is a crucial step in EEG signal processing, aimed at identifying and selecting the most informative and relevant features from high-dimensional datasets. Given the high noise levels and large volumes of EEG data, effective feature selection is essential for accurately interpreting brain activity. Techniques range from traditional statistical tests like the t-test and analysis of variance to more complex methods such as wrapper and embedded techniques \cite{t,anova}. Innovations in DL, including autoencoders and tree-based models, further improve feature selection by automatically identifying and quantifying the importance of EEG features \cite{select-dl}. For a detailed overview, refer to previous reviews \cite{select1, signalpro}.

\subsection{Machine Learning of Single-channel EEG}
In the processing of single-channel EEG signals, ML has significantly advanced automation and precision in targeted downstream tasks \cite{craik2019deep}. Traditional ML methods typically involve manual extraction and selection of features such as time domain statistical features, frequency domain features, and time-frequency domain features. These features are subsequently utilized to train various ML models like support vector machines (SVM), random forests (RF), and k-nearest neighbors (KNN) for specific tasks. While these models have demonstrated effective performance in EEG signal classification, their accuracy heavily relies on the quality and relevance of the chosen features \cite{chung,khatun,jalemotion}. 

The DL techniques have shown significant efficacy in EEG signal processing through multi-layer nonlinear architectures that automatically learn intricate data features and derive complex feature representations without manual engineering \cite{TinySleepNet,ay}, unlike traditional ML methods. For instance, convolutional neural networks (CNNs) excel in processing one-dimensional sequence data and have been successfully applied to automatic EEG feature extraction \cite{Anattention-based}. Recurrent neural networks (RNNs) and long short-term memory networks (LSTMs) effectively handle the time series characteristics of EEG signals and capture temporal dependencies \cite{TinySleepNet}. These capabilities have enabled DL models to achieve significant progress in tasks such as EEG signal classification, anomaly detection, and pattern recognition. Although DL excels in feature learning, it typically requires extensive training data and has poor interpretability. Conversely, traditional ML methods can operate with smaller datasets and offer more transparent decision-making processes. Therefore, selecting an appropriate ML strategy in practical applications should consider data availability, task complexity, and the need for model interpretability.

\section{APPLICATIONS OF SINGLE-CHANNEL EEG}
\subsection{Sleep Staging}
Sleep staging is a crucial aspect of sleep medicine and research, aiding in the understanding of different sleep stages \cite{U-Sleep}. According to the American Association of Sleep Medicine (AASM) scoring manual, sleep is divided into five stages: non-REM sleep (NREM, including N1, N2, N3), REM sleep (REM), and Wake (W) \cite{AASM}. Accurate sleep stage scoring provides valuable clinical insights. Polysomnography (PSG), the gold standard for diagnosing sleep disorders, involves monitoring EEG, ECG, EOG, EMG, airflow, thoracic and abdominal movements, and oximetry \cite{PSG}. Traditionally, sleep stages are classified by doctors based on these physiological signals \cite{Europeansleepcenters}. This process typically involves recording at least two EEG channels, EOG, and EMG, and manually scoring them using Rechtschaffen and Kales rules \cite{AASMandRechtschaffen}. Single-channel EEG offers a viable alternative, capable of long-term all-night sleep measurement with minimal physiological burden and providing accurate, reliable data \cite{sleepcurrenttrends, Q-factor}. These attributes make it suitable for clinical research and home sleep monitoring. Sleep staging primarily relies on EEG frequency and waveforms \cite{mocks1984select}. While PSG systems used in sleep research provide multiple EEG channels, in practice, only one channel is often selected for analysis based on signal quality, anatomical location, and prior study support.

The Fpz-Cz channel has been extensively utilized in sleep studies due to its efficiency in detecting wakefulness and REM stages, as well as its ease of use \cite{Automaticsleepstaging, SleepEEGNet, Khalili}. Wei et al. employed Fpz-Cz channel EEG data from the Sleep-EDFx database, combining contextual scale maps and Co-ScaleNet to achieve high accuracy and recall in automatic sleep staging \cite{Automaticsleepstaging}. Their method proved robust across multiple datasets, including those from healthy individuals, patients with depression and other public datasets, particularly in the N1 stage. The semi-supervised learning model SHNN was proposed for automatic sleep staging using Fpz-Cz channel EEG data from the Sleep-EDFx dataset. This model demonstrated high classification accuracy and consistency, especially in light and deep sleep stages \cite{SHNN}. Studies by \cite{SleepContextNet, SingleChannelNet} used single-channel EEG data from the Fpz-Cz channel for automated sleep staging, exploring sleep staging directly from EEG signals using temporal context information. Eldele et al. proposed an attention-based DL architecture, AttnSleep, to classify sleep stages using single-channel EEG signals, whose architecture relies on a multi-resolution CNN and adaptive feature recalibration \cite{Anattention-based}. An efficient single-channel EEG analysis model, designed for sleep stage scoring on the Fpz-Cz channel and based on CNN and LSTM, was developed in \cite{TinySleepNet}. An automatic sleep staging system leveraging local extreme value statistical behavior for feature extraction and a residual attention model-based method for accurate EEG sleep staging have been developed, both focusing on the Fpz-Cz channel \cite{Anewautomatic, Aresidualbased}.

The C4-A1 and Cz-A1 channels have demonstrated effective performance for sleep staging tasks on certain datasets. In the SHHS dataset, the C4-A1 channel shows a clear advantage by capturing more information when processing complex EEG signals, thereby improving classification accuracy \cite{SleepContextNet, Endtoend}. The Cz-A1 channel, known for providing stable signals, helps reduce misclassification due to eye movements and its application across multiple datasets has yielded consistent and high-precision classification results. The local extreme value statistical method applied on the C4-A1 channel validated its effectiveness for automatic sleep staging \cite{Anewautomatic}, while subsequent experiments with the Cz-A1 channel demonstrated excellent classification performance and advantages in signal stability and eye movement interference reduction \cite{Anattention-based}. Furthermore, Supratak et al. proposed the TinySleepNet model, which employs DL technology for efficient single-channel EEG analysis. Their study utilized the Cz-A1 channel to verify its efficiency and versatility across multiple datasets. The TinySleepNet model achieves performance comparable to or better than existing methods while reducing the number of parameters and computing resources required \cite{TinySleepNet}.

In addition to the Fpz-Cz, Cz-A1, and C4-A1 channels, other studies have also selected channels such as C3-A2 and Pz-Oz \cite{Automaticsleepstaging, Aresidualbased, Adecisionsupport}. Different EEG channels exhibit unique advantages and application scenarios in sleep staging, which are summarized in Table \ref{table3}.

\renewcommand\arraystretch{1.5}
\begin{table*}
\centering
\caption{Article summary of sleep staging based on single-channel EEG systems}

 \begin{tabular}{p{2.7cm}<{\centering}p{1.0cm}<{\centering}p{3.8cm}<{\centering}p{1.4cm}<{\centering}p{1.1cm}<{\centering}p{1.2cm}<{\centering}p{1cm}<{\centering}p{1.2cm}<{\centering}}
 
  \Xhline{1.2pt}
        \textbf{Dataset} & \textbf{Ref} & \textbf{Sleep Stages Classified} & \textbf{Channel} & \textbf{Accuracy(\%)} & \textbf{Kappa} & \textbf{Validation} & \textbf{Subjects} \\ 
        \Xhline{1.2pt}
\textbf{Sleep-EDFx \cite{Sleep-EDFx}} & \cite{Automaticsleepstaging}   & Wake-N1-N2-N3-REM    & Fpz-Cz & 82.8  & 0.77 & PSG & 153 \\
           & \cite{SleepEEGNet}    & Wake-N1-N2-N3-REM  & Fpz-Cz & 80.3  & 0.73 & PSG & 78 \\
           & \cite{Khalili}                                      & Wake-N1-N2-N3-REM    & Fpz-Cz & 82.5  & 0.76 & PSG & 78  \\
           & \cite{lv}                                      & Wake-N1-N2-N3-REM    & Fpz-Cz & 81.0  & 0.74 & PSG & 78 \\
           & \cite{huang}                                   & Wake-N1-N2-N3-REM    & Fpz-Cz & 82.1  & 0.79 & PSG & 78  \\
           & \multirow{2}{*}{\cite{SHNN}}                    & Wake-N1-N2-N3-REM    & Fpz-Cz & 78.6  & 0.71 & PSG & 153 \\
           &                                          & Wake-N1-N2-N3-REM    & Pz-Oz  & 73.4  & 0.63 & PSG & 153 \\
           & \cite{SleepContextNet}                         & Wake-N1-N2-N3-REM    & FPz-Cz & 82.8  & 0.76 & PSG & 78  \\
           & \cite{SingleChannelNet}                       & Wake-N1-N2-N3-REM    & FPz-Cz & 86.1  & 0.81 & PSG & 78  \\
           & \cite{Endtoend}                                & Wake-N1-N2-N3-REM    & FPz-Cz & 81.3  & -    & PSG & 100 \\
           & \cite{Anattention-based}                        & Wake-N1-N2-N3-REM    & FPz-Cz & 81.3  & 0.74 & PSG & 78  \\
           & \cite{TinySleepNet}                             & Wake-N1-N2-N3-REM    & FPz-Cz & 85.4  & 0.80 & PSG & 78  \\
  \Xhline{1.2pt}
\textbf{ISRUC \cite{ISRUC}}     & \cite{shen}                                     & Wake-N1-N2-N3-REM    & C3-A2  & 81.65 & 0.76 & PSG & 100 \\
           & \cite{Gharbali}                                 & Wake-N1-N2-N3-REM    & C3-A2  & 84.0  & -    & PSG & 100 \\
           & \cite{Automaticsleepstaging}                    & Wake-N1-N2-N3-REM    & C3-A2  & 84.6  & 0.80 & PSG & 100 \\
           & \cite{AutomaticSleepStage}                      & Wake-N1-N2-N3-REM    & F3-A2  & 80.4  & 0.75 & PSG & 100 \\
  \Xhline{1.2pt}
\textbf{DREAMS-SUB \cite{DREAMS-SUB}} & \cite{SHNN}                                    & Wake-N1-N2-N3-REM    & Cz-A1  & 78.2  & 0.69 & PSG & 20  \\
           & \cite{Automaticsleepstaging}                    & Wake-N1-N2-N3-REM    & Cz-A1  & 83.4  & 0.77 & PSG & 20  \\
           & \cite{Anewautomatic}                            & Wake-S1-S2-S3-S4-REM & Cz-A1  & 83.4  & 0.77 & PSG & 20  \\
           & \cite{Automatedidentification}                  & Wake-S1-S2-S3-S4-REM & Cz-A1  & 70.7  & 0.71 & PSG & 20  \\
           & \cite{Adecisionsupport}                         & Wake-N1-N2-N3-REM    & Cz-A1  & 78.9  & 0.78 & PSG & 20  \\
  \Xhline{1.2pt}
\textbf{SHHS \cite{SHHS}}       & \cite{Li}     & Wake-N1-N2-N3-REM    & C4-A1  & 85.1  & 0.79 & PSG & 329 \\
           & \cite{Automaticsleepstaging} & Wake-N1-N2-N3-REM    & C4-A1  & 87.9  & 0.83 & PSG & 329 \\
           & \cite{SleepContextNet} & Wake-N1-N2-N3-REM    & C4-A1  & 86.4  & 0.81 & PSG & 329 \\
           & \cite{Automaticsleepstaging}  & Wake-N1-N2-N3-REM    & C4-A1  & 85.8  & 0.80 & PSG & 111 \\
           & \cite{Endtoend}    & Wake-N1-N2-N3-REM    & C4-A1  & 86.7  & - & PSG & 100 \\
           & \cite{Anattention-based} & Wake-N1-N2-N3-REM    & C4-A1  & 84.2  & 0.78 & PSG & 329 \\
  \Xhline{1.2pt}
\textbf{Sleep-EDF \cite{Sleep-EDF}}  & \cite{SHNN}  & Wake-N1-N2-N3-REM    & FPz-Cz & 84.8  & 0.79 & PSG & 20  \\
           & \cite{SingleChannelNet} & Wake-N1-N2-N3-REM    & FPz-Cz & 91.0  & 0.88 & PSG & 20  \\
           & \cite{Automaticsleepstaging} & Wake-N1-N2-N3-REM    & Pz-Oz  & 91.9  & 0.87 & PSG & 8   \\
           & \multirow{2}{*}{\cite{CCRRSleepNet}}            & Wake-N1-N2-N3-REM    & FPz-Cz & 84.3  & 0.78 & PSG & 20  \\
           &                                          & Wake-N1-N2-N3-REM    & Pz-Oz  & 80.31 & 0.73 & PSG & 20  \\
           & \cite{Anattention-based}                        & Wake-N1-N2-N3-REM    & FPz-Cz & 84.4  & 0.79 & PSG & 20  \\
           & \multirow{2}{*}{\cite{Aresidualbased}}          & Wake-N1-N2-N3-REM    & FPz-Cz & 84.3  & 0.78 & PSG & 20  \\
           &                                          & Wake-N1-N2-N3-REM    & Pz-Oz  & 80.7  & 0.74 & PSG & 20  \\
           & \cite{TinySleepNet}                             & Wake-N1-N2-N3-REM    & FPz-Cz & 83.1  & 0.77 & PSG & 20  \\
           & \multirow{2}{*}{\cite{Deepconvolutionalneural}} & Wake-S1-S2-S3-S4-REM & Pz-Oz  & 93.6  & -    & PSG & 20  \\
           &                                          & Wake-S1-S2-S3-S4-REM & FPz-Cz & 91.1  & -    & PSG & 20  \\
           & \multirow{2}{*}{\cite{Anewautomatic}}           & Wake-S1-S2-S3-S4-REM & FPz-Cz & 90.6  & 0.85 & PSG & 20  \\
           &                                          & Wake-S1-S2-S3-S4-REM & Pz-Oz  & 88.6  & 0.82 & PSG & 20  \\
           & \multirow{2}{*}{\cite{DeepSleepNet}}            & Wake-N1-N2-N3-REM    & FPz-Cz & 82.0  & 0.76 & PSG & 20  \\
           &                                          & Wake-N1-N2-N3-REM    & Pz-Oz  & 79.8  & 0.72 & PSG & 20  \\
           & \cite{Anautomatedmethod}                      & Wake-S1-S2-S3-S4-REM & Pz-Oz  & 90.0  & -    & PSG & 20  \\
           & \cite{Single-channelEEGsleepstage}              & Wake-S1-S2-S3-S4-REM & FPz-Cz & 90.5  & 0.80 & PSG & 20  \\
           & \cite{Automatedidentification}                  & Wake-S1-S2-S3-S4-REM & Pz-Oz  & 88.1  & 0.88 & PSG & 20  \\
           & \cite{Adecisionsupport}                         & Wake-N1-N2-N3-REM    & Pz-Oz  & 93.7  & 0.85 & PSG & 20  \\
    \Xhline{1.2pt}
  \end{tabular}
    \label{table3}
\end{table*}

\subsection{Emotion Recognition}
\begin{figure}
    \centering
    \includegraphics[width = 2.8in]{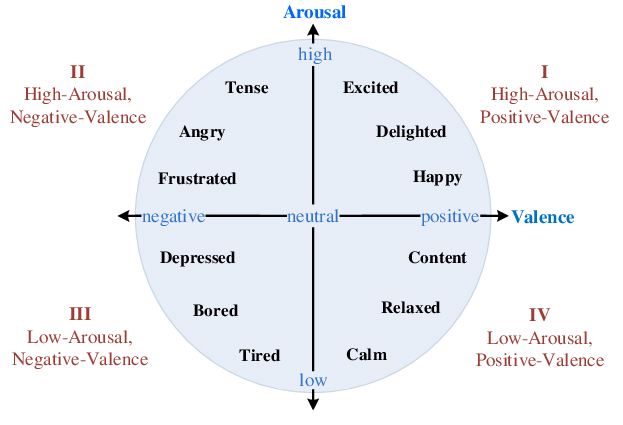}
    \caption{The Arousal–Valence Space (AVS) Model used in psychology and affective computing to represent and measure human emotions \cite{AVS}.}
    \label{AVS}
\end{figure}

Emotion is a multifaceted psychological and physiological state observed in humans and other organisms in response to external or internal stimuli. Emotions entail internal physiological changes including variations in heart rate, respiration, and brain electrical activity \cite{emotioneeg}. Arousal level denotes the extent of physiological and psychological activation experienced during an emotion, ranging from negative to positive states \cite{arousal}. Valence refers to the pleasantness of an emotion, indicating whether it is positive or negative and is also measured on a negative to positive scale \cite{valence}. Integrating arousal level and valence allows for a comprehensive classification of emotional states \cite{AVS}, as illustrated in Fig. \ref{AVS}, providing a clear framework for emotion recognition and regulation. 

With the advent of wireless head-mounted single-channel EEG devices, the requirement for conductive glue has diminished, enhancing their portability and flexibility. This development has positioned them as the primary option for emotion recognition research, owing to their ease of use and mobility \cite{emotion2}. Table \ref{emotion} presents various applications of single-channel EEG in emotion recognition. Zhao et al. introduced an interpretable method for emotion classification using single-channel EEG signals based on the Emo\_Food dataset \cite{Emo-Food}. They employed wavelet packet decomposition and statistical feature extraction for multi-domain features, using extreme gradient boosting (XGBoost) \cite{xgboost} and Shapley additive explanation (SHAP) method for interpretable emotion classification \cite{Emo-Food}. Yan et al. proposed an emotion recognition method using single-channel EEG signals from DEAP \cite{deap} and DREAMER datasets \cite{dreamer}. They employed topological nonlinear dynamic analysis, including phase space reconstruction and persistent homology, to study EEG characteristics during emotional states \cite{yan}. Khan et al. investigated emotion induction through video games using single-channel EEG equipment. Participants across various age groups played different games to induce emotions, and EEG signals were analyzed alongside sentiment classification using ML techniques \cite{khan}. Cao et al. developed an emotion recognition method tailored for single-channel EEG signals. Their approach involved WT for signal decomposition, extraction of time-frequency information from multiple frequency bands, and sliding time windows for feature calculation and emotion classification \cite{caoemotion}. Jalilifard et al. studied the classification of emotional states using single-channel EEG, focusing on the left prefrontal cortex (Fp1 location). They utilized emotional stimulus videos and applied KNN and SVM algorithms for successful classification of fear and relaxation emotions \cite{jalemotion}. Acharya et al. proposed a genetic programming (GP) framework with a new fitness function {``}Gap Score (G Score)" for imbalanced emotion recognition datasets. They used data collected from the NeuroSky MindWave II single-channel EEG device to classify four emotion types: happy, scary, sad, and neutral \cite{acharyaemotion}. Li et al. proposed a brain rhythm sequencing (BRS) technique with dynamic time warping for single-channel selection, achieving 70-82\% classification accuracy using 10-second brain rhythm sequences for emotion recognition \cite{liemotion}.

These findings demonstrate that single-channel EEG achieves high accuracy and effectiveness in emotion classification, highlighting the potential of single-channel EEG devices in future affective computing and emotion recognition.

\subsection{Neurofeedback}
\begin{table*}[!htbp] 
	\centering
	\caption{Article summary of Neurofeedback based on Single-channel EEG Systems}
	
	\begin{tabular}{p{1cm}<{\centering}p{3cm}<{\centering}p{2cm}<{\centering}p{2cm}<{\centering}p{1.5cm}<{\centering}p{4.5cm}<{\centering}} 
		\Xhline{1.2pt} 
		
		\textbf{Ref} & \textbf{Device} & \textbf{Selected Channel} & \textbf{Frequency Band} & \textbf{Subject Number} & \textbf{Application Domain} \\ 
		\Xhline{1.2pt}
		\cite{sultanov2023mobile}   & NeuroSky MindWave Mobile & Fp1  & $\alpha$ & 16 & Anxiety  \\ \hline
\cite{israsena2021brain}   & Neurosky EEG headset & Fp1  & high $\alpha$, low $\alpha$, $\beta$, low $\beta$, $\theta$ & 40    &   Enchancing cognitive performance                                                                                                                                                                                 \\ \hline
\cite{toth2023developing}   & Self-built Device    & Fpz        & \begin{tabular}[c]{@{}l@{}}$\theta$, $\alpha$, and $\beta$\end{tabular} & 6              & Home nursing               \\                                                                                                    
		\Xhline{1.2pt} 
	\end{tabular}
	\label{neur}
\end{table*}
Neurofeedback is a safe, non-invasive technique grounded in learning theories from neuroscience and behavioral science, aimed at enhancing brain function and structure. By measuring neural activity associated with specific functions and delivering real-time feedback through visual, auditory, or tactile stimuli, neurofeedback encourages individuals to employ psychological strategies to voluntarily modulate targeted neural activity, thereby enhancing behavior and cognitive function. Neurofeedback has demonstrated positive outcomes in managing depression and anxiety \cite{hammond2005neurofeedback}. However, limited training sessions, small sample sizes, lack of standardization and unclear long-term effects constrain its efficacy, requiring further research to address these limitations \cite{omejc2019review}.

A comprehensive investigation with NeuroSky's wireless fronto-polar EEG monitoring revealed significant inverse correlation between $\alpha$ wave activity and anxiety levels ($p$ $\leq$ 0.029) in adolescent athletes during eyes-closed conditions, demonstrating single-channel EEG systems' efficacy for psychological assessment and athletic anxiety management \cite{sultanov2023mobile}. A cognitive intervention study utilizing cost-effective single-channel EEG devices \cite{israsena2021brain} implemented five brain-training protocols across 20 sessions in five Thai medical centers, demonstrating statistically significant enhancements ($p$ $\leq$ 0.05) in visual memory, attention, and recognition among cognitively-screened elderly participants, thus establishing single-channel EEG neurofeedback as a viable therapeutic intervention for cognitive enhancement in aging populations. Sandra et al. \cite{loo2012clinical} elucidate the clinical applications of EEG in attention deficit hyperactivity disorder (ADHD) research and treatment, emphasizing EEG-based neurofeedback therapeutic efficacy while acknowledging diagnostic heterogeneity. Hande et al.'s \cite{kaynak2019effect} investigation implemented neurofeedback training protocols targeting $\theta$ and sensorimotor rhythm (12-15 Hz) bands in adult cohorts, evaluating selective attention through stroop task measurements.

Extant neurofeedback research faces methodological limitations from insufficient training sessions and suboptimal task complexity, requiring enhanced protocols with increased sessions and challenging paradigms for conclusive clinical efficacy evidence, as summarized in Table \ref{neur} for single-channel EEG interventions.

\subsection{Educational Research}
Single-channel EEG monitors brain activity changes to assess cognitive load during learning tasks and stages, enabling optimization of teaching content, methods and outcomes \cite{education-review1}. In classroom environments, it monitors attention changes to teaching content, helping teachers improve strategies and rhythm. When evaluating educational technology and online learning tools, single-channel EEG provides objective evaluation metrics by measuring changes in brain activity with different tools.

Research examining single-channel EEG in educational contexts has yielded diverse findings. Initial investigations by Wei et al. utilized NeuroSky MindBand\cite{MindwaveMobile} to study visual attention and reading duration effects on children and adults during picture book reading \cite{weima}. Subsequent mobile environment research assessed learner attention variations across sitting, standing and walking scenarios through NeuroSky MindSet headset \cite{chenlin}. Further studies investigated speech-to-text applications' impact on academic performance, attention and meditation through NeuroSky MindWave \cite{wuhuang, MindWave}. Additional research explored children's attention during reading across varied media formats including traditional, pop-up, audio and e-books, focusing on grades 3-6 while analyzing gender and grade-level influences \cite{mawei}. Research examining video lecture methods analyzed how different presentation styles affected student sustained attention \cite{chenwu}, while parallel studies investigated attention state changes under varying text display types \cite{wanghsu}.

Studies in traditional classroom settings demonstrated that rewarding correct answers with interesting images effectively stimulated positive student emotions \cite{huangliu}. Comprehensive research in intelligent tutoring systems compared clicker technology with mobile voting technology, examining effects on student anxiety, self-efficacy, engagement and academic performance \cite{inventado}. Additional investigations measured attention levels across varying difficulty levels in computer-based learning \cite{wanghsu} and online synchronous teaching activities \cite{chenwang, ghe, assess}. Research in motor skill acquisition investigated attention during computerized visual motor tasks \cite{wongchan}, while studies in emotional learning systems examined brainwave concentration and relaxation training, incorporating automatic material adjustment based on emotional states \cite{lin}.

\begin{table*}[!htbp] 
\centering
\caption{Article summary of Emotion recognition based on Single-channel EEG Systems}

\begin{tabular}{p{0.5cm}<{\centering}p{1.8cm}<{\centering}p{1.5cm}<{\centering}p{1.1cm}<{\centering}p{3.3cm}<{\centering}p{1.5cm}<{\centering}p{1.2cm}<{\centering}p{2.8cm}<{\centering}p{1cm}<{\centering}} 
  \Xhline{1.2pt} 
   
\textbf{Ref} & \textbf{Device} & \textbf{Datasets} & \textbf{Selected Channel} & \textbf{Features} & \textbf{Model} & \textbf{Emotion Classes} & \textbf{Emotion} & \textbf{Result(\%)} \\ 
\Xhline{1.2pt}
\cite{Emo-Food} & Neurosky Mindwave Mobile II & Emo\_Food Dataset & Fp1 & Four frequency bands($\alpha$,low $\beta$,high $\beta$, $\gamma$) & XGBoost & 3 & Neutral, Positive, Negative & 92.31 \\ \hline

\multirow{4}{*}{ \cite{liemotion} } & \multirow{4}{*}{\begin{tabular}[c]{@{}l@{}}Multi-channel \\ system device\end{tabular}} & MER Dataset        &      & $\delta$, $\theta$, $\alpha$, $\beta$, and $\gamma$ & BRS & Continuous & 5 emotional factors: familiarity, liking, understanding, arousal, and valence(1-9)  & 72.11  \\
                    &   & SEED Dataset       &  & $\delta$, $\theta$, $\alpha$, $\beta$, and $\gamma$ & BRS & 3 & Neutral, Positive, Negative   & 72.40  \\
                    &  & DEAP Dataset       & C3     & $\delta$, $\theta$, $\alpha$, $\beta$, and $\gamma$ & BRS         & Continuous & 2 emotional factors: arousal, valence(1-9)             & 74.51  \\
                    &  & MAHNOB Dataset     &     & $\delta$, $\theta$, $\alpha$, $\beta$, and $\gamma$ & BRS & Continuous & 2 emotional factors: arousal, valence(1-9)  & 82.18  \\ \hline
		\cite{khan} & Neurosky Mindwave Mobile II & Self-built Dataset & Fp1 & Time, Frequency and Time-frequency Domains & Boosted Trees classifier & 4 & Happy, Bored, Relaxed, Stressed & 82.26 \\ \hline

\cite{caoemotion} & Neurosky Mindwave & Emo\_Food Dataset & Fp1 & Four frequency bands($\alpha$,low $\beta$,high $\beta$, $\gamma$) & TF-LSTM & 3 & Neutral, Positive, Negative & 98.36 \\ \hline

\cite{jalemotion} & Neurosky Mindwave Mobile & Self-built Dataset & Fp1 & Four frequency bands($\theta$, $\alpha$, $\beta$, $\gamma$) & KNN & 3 & Neutral, Relaxation, Scary & 94.21 \\ \hline

\cite{acharyaemotion} & Neurosky Mindwave Mobile II & Self-built Dataset & Fp1 & Attention, Meditation, $\delta$, $\theta$, Low $\alpha$, High $\alpha$, Low $\beta$, High $\beta$, Low $\gamma$, and High $\gamma$ & GP & 4 & Happy, Sad, Horror, Neutral & 87.61 \\ \hline
  \Xhline{1.2pt} 
  \end{tabular}
\begin{tablenotes}
 \item[] \textbf{XGBoost}: Extreme Gradient Boosting. \textbf{BRS}: Brain Rhythm Sequences. \textbf{TF-LSTM}: Time-Frequency Long Short-Term Memory. \textbf{KNN}: K-Nearest Neighbors. \textbf{GP}: Genetic Programming. \textbf{Continuous}: measured on a continuous scale, not discrete.
\end{tablenotes}
    \label{emotion}
\end{table*}

\subsection{Clinical Diagnosis and Treatment Assessment}
Single-channel EEG has gained significance in medical diagnosis through swift and effective brain state assessments. It is particularly crucial for rapid decision-making in acute conditions like stroke onset and epileptic seizures. In mental health conditions, single-channel EEG enables continuous monitoring through detectable pattern changes, supporting clinical diagnosis and treatment evaluation. For neurodegenerative conditions like Alzheimer's Disease (AD) \cite{ad} and Mild Cognitive Impairment (MCI) \cite{mci}, single-channel EEG detects brain wave anomalies indicative of cognitive decline, enabling early detection and intervention. Its simplicity and accessibility position single-channel EEG as a promising tool for large-scale screening.

\subsubsection{Depression}
Single-channel EEG offers a noteworthy advantage due to its simplicity and rapid deployment capability, facilitating timely and potentially more frequent assessments in mental health contexts. Depression, a prevalent and debilitating disorder, necessitates diagnostic tools that are effective, accessible, and efficient. Studies using single-channel EEG for diagnosing depression have generally focused on analyzing brain wave patterns and connectivity measures to distinguish between depressed patients and healthy controls \cite{hasanzadeh, chung}. Much of this research aims to identify specific EEG biomarkers indicative of depression. For instance, alterations in frontal $\alpha$ asymmetry have been commonly observed in depressive patients, suggesting a potential diagnostic signature detectable through single-channel EEG \cite{alpha_alter}. A pivotal study demonstrated that single-channel EEG analysis of frontal electrode signals through ML techniques effectively identifies depression via reduced left frontal activity, achieving diagnostic accuracy comparable to multi-channel systems \cite{chung}. Another approach examines EEG responses to cognitive or emotional stimuli tasks, analyzing electrical brain response differences between depressed patients and controls to reveal neural mechanisms underlying emotional dysregulation \cite{hasanzadeh}. Depression studies investigate various EEG features including spectral power across frequency bands ($\alpha$, $\beta$, $\gamma$, $\delta$, $\theta$), asymmetry indices, and coherence measures, with particular emphasis on the alpha asymmetry index as a potential depression biomarker through assessment of alpha wave activity disparity between hemispheres, where greater left-sided inactivity correlates with depressive symptoms \cite{alpha}. 

Chung et al. proposed a method for assessing depression using a single-channel EEG device (NeuroSky Mindwave Mobile II), demonstrating efficient assessment through multi-layer ensemble learning technology \cite{chung}. Zhu et al. introduced a method for detecting depression through single-channel EEG and graph-based techniques in 2022, achieving highest accuracy of 92.0\% with the T4 channel during eyes-closed states \cite{zhu}. Shen et al. presented an enhanced channel selection approach using kernel target alignment for depression detection in multi-channel EEG, showing improved classification performance and satisfactory results with a single channel aligned with cortical activity patterns associated with depression \cite{shendepre}. Wan et al. explored a ML method for distinguishing major depressive disorder (MDD) from normal controls using single-channel EEG data from forehead locations (Fp1 and Fp2 electrodes). Their results indicated superior classification performance at Fp1, showcasing single-channel EEG's capability comparable to multi-channel EEG in MDD differentiation \cite{wan}. Ay et al. proposed a deep hybrid model combining CNN and LSTM architecture to detect depression through EEG signals, achieving high classification accuracy of EEG signals in the right and left hemispheres 99.12\% and 97.66\%, respectively \cite{ay}. Additional studies \cite{ac,ba1,ba2} underscored the effectiveness of single-channel EEG analysis across various methods for detecting depression, emphasizing its feasibility and efficacy in linear and nonlinear EEG signal analysis techniques.

\subsubsection{AD \& MCI}
Alzheimer’s Disease and Mild Cognitive Impairment represent a continuum of cognitive decline that poses significant diagnostic challenges. Early detection of MCI, especially its differentiation from age-related cognitive changes, is crucial as MCI can often progress to AD. Single-channel EEG, with its simplicity and cost-effectiveness, offers a promising tool for early diagnosis and monitoring of these conditions. Khatun et al. employed the single-channel EEG to analyze brain responses to speech stimulation for MCI detection, which achieved up to 87.9\% cross-validation accuracy, 85\% sensitivity, and 90\% specificity, demonstrating the potential for early MCI detection using single-channel EEG and event-related potentials (ERP) \cite{khatun}. Pulver et al. used single-channel sleep EEG to analyze 205 elderly adults, exploring changes in rhythmic events such as theta wave bursts, sleep spirals, and slow waves generated by memory-related neural circuits during the early pathological process of AD \cite{pulver}. Mitsukura et al. utilized a single-channel EEG device \cite{MindwaveMobile} to record data from 120 patients with dementia and MCI, compared to healthy controls. The study found significant power spectrum differences between dementia patients and healthy controls at 3 Hz, 4 Hz, 10 Hz, and higher frequencies, as well as differences between patients with varying disease severity and healthy individuals \cite{mit}. These studies imply the validity of EEG as a biomarker for MCI and dementia in clinical practice, but further research with larger sample sizes is needed to verify its reproducibility. 

\subsubsection{Epilepsy}
Epilepsy, characterized by recurrent and unprovoked seizures, represents one of the most prevalent neurological disorders globally \cite{epilepsy1}. While multi-channel EEG remains the standard diagnostic tool, its complexity, high cost, and specialized requirements limit widespread application, particularly in resource-limited settings. Single-channel EEG emerges as a simpler, cost-effective alternative with significant clinical utility. Dweiri et al. created a new ML algorithm using XGBoost for seizure detection in wearable systems, classifying single-channel EEG from the CHB-MIT database, which demonstrated high sensitivity of 89\% for seizure detection \cite{yaz}. Another notable study placed sensors in varying numbers of channels to detect epilepsy separately using the CHB-MIT dataset. The results indicated that the single-channel method's performance in identifying epileptic seizures in continuous EEG recordings was comparable to the multi-channel method, with high sensitivity and a low false alarm rate \cite{chung2024}. Lu et al. compared the diagnostic efficacy of single-channel and standard multi-channel EEG in a clinical trial involving patients suspected of nocturnal seizures. Results suggested that while single-channel EEG is less detailed, it still correctly identified significant seizure activities in the majority of cases \cite{lu}. This finding supports the potential of single-channel EEG as a screening tool, particularly in environments where access to full EEG setups is not feasible.

\subsection{Other Applications}
The single-channel EEG with SSVEP paradigm demonstrates significant potential in BCI, neuroscience research and clinical diagnosis, with devices typically recording signals in the occipital region for visual stimuli sensitivity. Nguyen et al. developed a SSVEP speller system using monopolar single-channel EEG with one-dimensional CNN and a custom head-mounted device, achieving 99.2\% classification accuracy in experiments and 97.4\% accuracy with 49 ± 7.7 bits/min transmission rate in online spelling tests \cite{bipolar1}. Karunasena et al. demonstrated a streamlined SSVEP-based BCI system controlling a robotic arm through visual responses to frequency-specific LED stimuli \cite{Karunasena}.
The single-channel EEG with motor imagery paradigm is also a significant area of BCI research. Ge et al. used single-channel EEG to classify four types of motor imagery, decomposing the signal into the time-frequency domain through STFT, constructing multi-channel information, and combining the common spatial pattern algorithm and SVM to achieve high classification accuracy from the sensorimotor and frontal areas \cite{ge}.
Due to its user-friendliness, convenience, and low cost, single-channel EEG has also been applied in microsleep detection to enhance the safety of intelligent transportation systems (ITS). Chougule et al. developed Diamond Net, which uses an attention mechanism combined with WT and STFT spectrograms to effectively distinguish microsleep from other sleep stages \cite{its}.

\begin{figure}
	\centering
	\includegraphics[width = 2.5in]{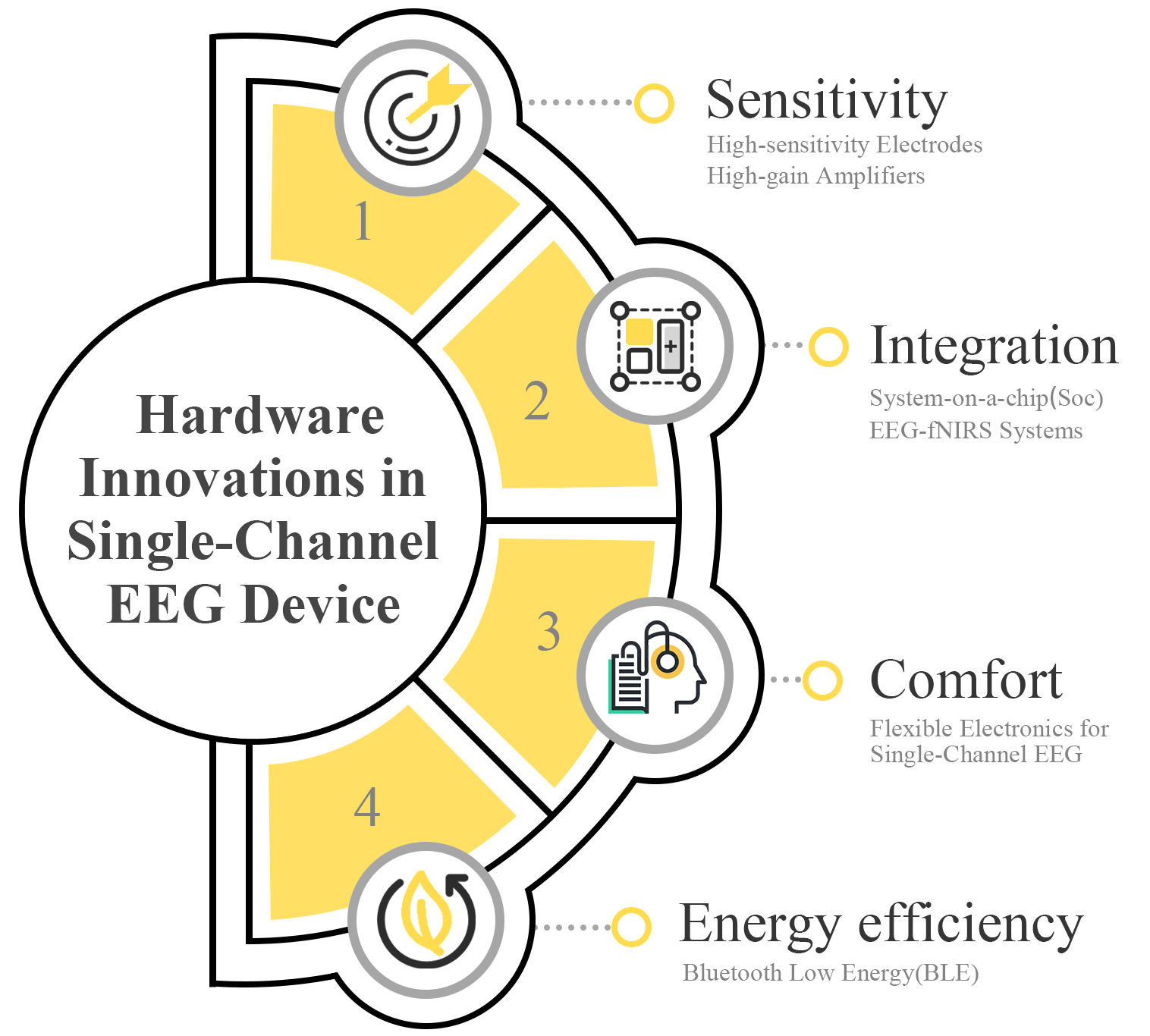}
	\caption{Advancements in single-channel EEG hardware focus on enhancing sensitivity, integration, comfort, and energy efficiency to improve device performance and user experience.}
	\label{hardware}
\end{figure}
\section{DISCUSSION}
Single-channel EEG demonstrates pivotal advantages through low cost, comfort, and non-invasiveness. The proliferation of research articles reflects its growing significance, driven by technological advancements toward accessible brain monitoring. Its adoption spans diverse applications from neuroscience exploration to clinical diagnosis, including human-computer interaction, education, and mental health monitoring. Firstly, a critical aspect of single-channel EEG's utility lies in its signal processing and data handling capabilities. Traditionally, EEG involved multiple channels for comprehensive brain activity mapping, but single-channel systems require sophisticated computational techniques to enhance accuracy and interpretability.
Secondly, single-channel EEG innovation emphasizes portability, reduced setup complexity, rapid deployment, and continuous monitoring capabilities. Advances in hardware and software through miniaturization, enhanced sensor sensitivity, and wireless capabilities promise improved signal quality and reliability, while sophisticated ML algorithms enable robust signal processing and data analysis.
Thirdly, significant challenges such as ensuring high signal quality and overcoming the inherent limitations of single-channel systems, like data depth and interpretative consistency, are expected to be further addressed through AI-based generative models. AI-based EEG generation techniques have demonstrated the potential to generate multi-channel signals with high quality from sparse-channel signals, matching or even surpassing the performance of multichannel systems \cite{chenhongyu}. 
Finally, rigorous validation and standardization of generative AI models in single-channel EEG systems are crucial, along with developing robust frameworks to navigate ethical and privacy concerns. Several critical areas of development are anticipated to influence the future trajectory of single-channel EEG technology.
\subsection{Advanced Signal Processing with AI Integration}

Recent advancements in signal processing and AI have transformed single-channel EEG applications through technological innovations: CNNs demonstrate exceptional spatial feature extraction capabilities \cite{rakhmatulin2024exploring}, one-dimensional CNNs excel in mental state classification and anomaly detection \cite{mental}, while LSTM networks effectively model temporal dependencies \cite{cascaded}, as evidenced by correlation-based epileptic signal classification methodology \cite{aliyu2023selection}. The field's progressive evolution encompasses sophisticated transfer learning strategies enabling cross-domain knowledge application in sleep staging scenarios \cite{he2023cross}, {advanced generative models including Generative Adversarial Networks (GANs) and Variational Autoencoders (VAEs) facilitating data augmentation and reconstruction} \cite{panwar2020modeling, EEG-GAN}, innovative auxiliary synthesis frameworks bridging the gap between single and multi-channel systems \cite{liang2023auxiliary}, refined adaptive filtering techniques for artifact mitigation in dynamic environments \cite{queiroz2022single, dhindsa2017filter}, and seamless edge computing integration enabling continuous monitoring and real-time neurofeedback applications \cite{nahavandi2022application}.

Future advancements in single-channel EEG technology will be fundamentally driven by specialized DL architectures optimized for single-channel data characteristics, enhancing both classification accuracy and computational efficiency. The evolution of transfer learning methodologies will enable effective adaptation of multi-channel-trained models to single-channel applications, incorporating advanced domain adaptation techniques \cite{wan2021review}. The integration of generative models will extend beyond multi-channel data reconstruction to enhance signal quality and generate synthetic training datasets. Concurrent development of real-time signal processing algorithms will facilitate sophisticated artifact rejection and noise reduction mechanisms for wearable devices, ensuring robust EEG measurements in dynamic environments \cite{lin2010review}. Moreover, advances in edge AI and on-device learning will empower wearable EEG systems with embedded intelligence, minimizing latency while preserving data privacy through reduced external transmission requirements \cite{merenda2020edge,bian2024device,pal2021epilepsy}. Collectively, these synergistic advancements will not only elevate the performance and applicability of single-channel EEG systems but also enhance their accessibility, thereby fostering broader adoption across clinical, consumer, and research domains.

\subsection{Technological Innovations in Hardware}

\begin{figure}
	\centering
	\includegraphics[width = 3in]{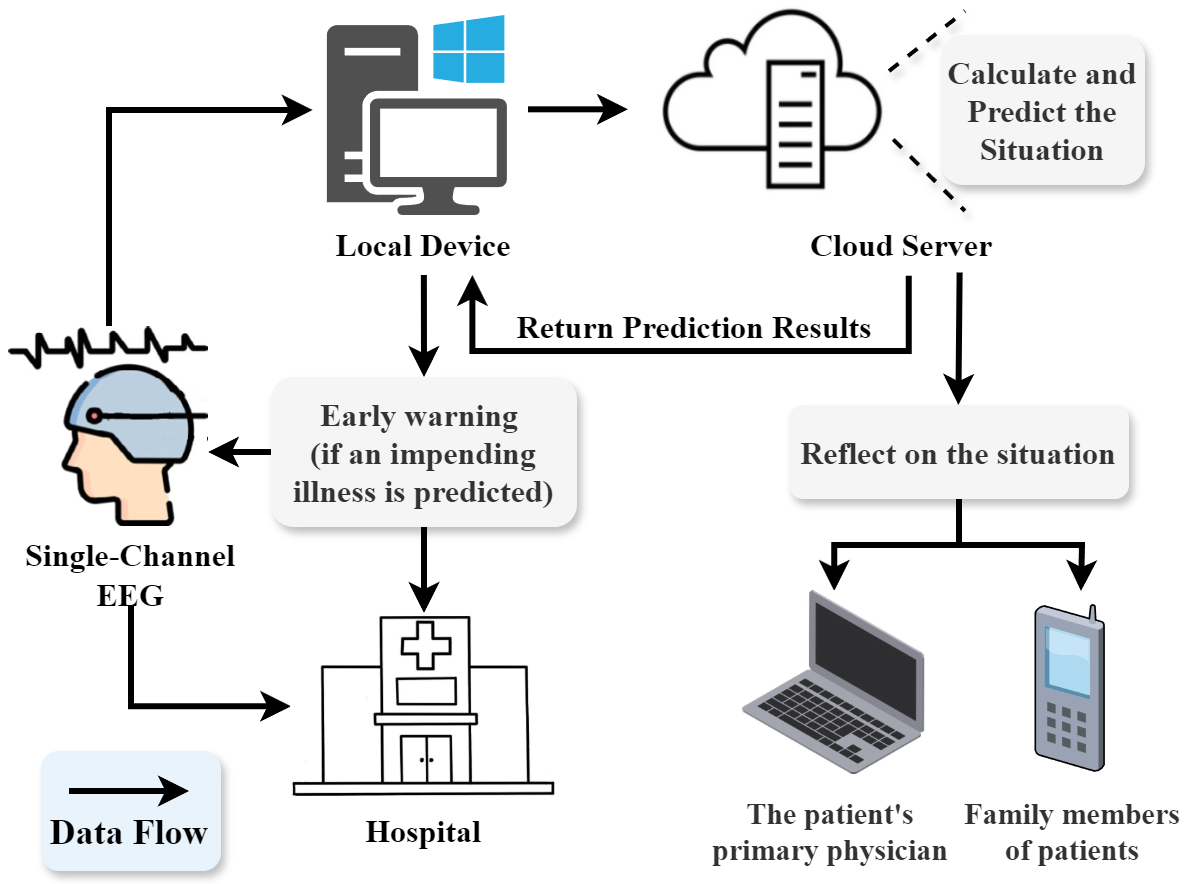}
	\caption{Integrating single-channel EEG data with wearable technology and IoT enables early disease prediction and alerts, facilitating timely intervention by sharing insights with physicians and family members.}
	\label{IoT}
\end{figure}
Hardware advancements have been fundamental to the evolution of single-channel EEG devices, with innovations primarily centered on four critical aspects: sensitivity, integration, comfort, and energy efficiency, as illustrated in Fig. \ref{hardware}. A typical EEG recording system comprises an amplifier, low pass filter, and ADC \cite{abbasi2021wearable}, where the amplifier serves as the crucial interface between electrode and signal processing engine. Contemporary high-gain amplifiers utilize low-noise architectures to achieve signal amplification while minimizing noise introduction. Marzieh et al. developed an innovative two-stage amplifier incorporating composite transistors and feedback structure, demonstrating superior direct current gain and slew rate, while achieving lower input reference noise compared to conventional metal-oxide-semiconductor (CMOS) amplifiers \cite{moradi2021designing}. Furthermore, Cheng et al. empirically validated that strategic selection of compensation capacitors significantly enhances amplifier performance, facilitating more precise brainwave signal acquisition \cite{cheng2023research}. Beyond amplifier innovations, electrode design has evolved significantly. Novel semi-dry and dry contact electrodes have emerged as promising alternatives to conventional Ag/AgCl wet electrodes \cite{li2020review}. Semi-dry electrodes combine advantages of wet and dry variants, delivering superior signal recording with minimal electrolyte usage, though their application remains limited to short-term monitoring due to electrolyte drying \cite{wang2023novel}. Dry electrodes excel in long-term monitoring while offering enhanced user comfort, particularly suitable for wireless wearable single-channel EEG devices. A critical frontier in electrode advancement lies in developing sophisticated materials to minimize contact impedance, with metals, carbon-based compounds, oxides, and polymers offering diverse fabrication possibilities \cite{yang2022materials}. Furthermore, the integration of flexible electronic technologies represents a pivotal research direction. While these technologies have predominantly been implemented in wet electrode systems \cite{shen2021novel}, their application in long-term dry electrode monitoring remains an active area of investigation, with numerous studies currently exploring this promising avenue \cite{luo2024fabrication}.

System on Chip (SoC) design represents a pivotal innovation in enhancing the integration of single-channel EEG devices. By consolidating multiple functional modules, such as signal acquisition, processing, and communication, onto a single chip, manufacturers can significantly reduce the device's footprint while improving overall performance. Mohammad et al. introduced a low-power, high-input-impedance EEG signal acquisition SoC that integrates an instrumentation amplifier, a neural signal-specific analog-to-digital converter (NSS-ADC), active electrodes, a transconductance-driven right leg, and a common-mode feedback module. This integration demonstrates its effectiveness and advantages in continuously acquiring EEG signals for wearable applications \cite{tohidi2019low}. Additionally, there is a growing interest in the integration of EEG with other modalities, particularly functional near-infrared spectroscopy (fNIRS), for a multimodal assessment of brain function. The EEG-fNIRS combination can provide a comprehensive picture of both electrophysiological and hemodynamic processes with high spatiotemporal resolution. The development of wearable, mechanically, and electrically integrated EEG-fNIRS technologies is a crucial next step in advancing this field \cite{uchitel2021wearable}. Furthermore, energy-efficient processors represent a significant innovation enabling prolonged EEG data collection and real-time analysis. Research advances in ultra-low-power technologies utilizing human body energy through heat or motion \cite{cao2024optimized} promise to eliminate battery dependency, advancing sustainable EEG monitoring capabilities..

\subsection{Integration with Wearable Technology and IoT}
The proliferation of wearable devices and Internet of Things (IoT) technology has revolutionized EEG signal acquisition, enabling more streamlined and naturalistic monitoring processes while facilitating intelligent automation through integrated detection systems \cite{iqbal2021advances}. Traditional wired EEG systems present significant limitations in terms of user experience and practical applicability \cite{he2023diversity}. In contrast, recent innovations in wireless EEG systems, particularly in wearable devices, enhance portability and user flexibility, advancing from conventional fixed systems toward intelligent wireless solutions prioritizing comfort and signal fidelity. While wearable devices present data quality and motion artifact challenges, their cost-effectiveness, wireless connectivity, lightweight construction, and operational simplicity, combined with continuous hardware improvements, demonstrate increasing potential and performance. Fig. \ref{IoT} demonstrates how the integration of single-channel EEG data with wearable technology and IoT infrastructure enables predictive disease detection and early warning systems, facilitating timely medical intervention through seamless information sharing among healthcare providers and family members.

The convergence of wearable devices and IoT technology facilitates specialized applications leveraging EEG monitoring, with detection systems demonstrating IoT device control capabilities \cite{laport2020prototype}. Single-channel EEG systems present a compelling integration trajectory through their minimal form factor and cost-effectiveness. While conventional multi-channel EEG systems deliver comprehensive neurological data, their substantial footprint limits practical deployment. Despite reduced spatial resolution, single-channel EEG devices employ advanced signal processing and algorithms for substantive neurophysiological information extraction, enabling mobile applications in real-time monitoring and personalized IoT environments. Research demonstrates applications in driver fatigue monitoring \cite{ju2024survey} and IoT-integrated seizure prediction through cloud services \cite{hosseini2017optimized}, implementing DL algorithms for prediction and automated healthcare notifications. IoT infrastructure capabilities \cite{ajmeria2022critical} with centralized cloud analytics enable efficient detection and rapid emergency response protocols. As technological capabilities evolve, single-channel EEG systems are projected to enhance the accessibility of neurological monitoring within IoT-enabled environments.

\section{CONCLUSION}
Single-channel EEG, benefiting from its low cost, comfort, and non-invasiveness, holds great potential for real-time brain activity monitoring. The increasing number of research articles underscores its growing significance. {This comprehensive review has analyzed the evolution and landscape of single-channel EEG, revealing advances in hardware design, signal processing methodologies, and applications. Consumer-grade devices have democratized brain-computer interaction, while sophisticated signal processing and ML algorithms have enhanced signal quality and feature extraction. These improvements have enabled robust applications across sleep analysis, emotion recognition, and clinical diagnostics, particularly for neurological conditions such as depression, AD, MCI, and epilepsy. AI integration has further expanded these capabilities, enabling more accurate classification and real-time analysis. The convergence of advanced hardware, signal processing, and AI presents opportunities for innovation, while future developments in AI-enhanced processing, smart hardware, and IoT connectivity will drive the next generation of applications, advancing the accessibility and reliability of brain monitoring technologies. We believe this review of single-channel EEG will facilitate future research and innovation in brain monitoring systems.}

\ifCLASSOPTIONcaptionsoff
  \newpage
\fi

\normalem
\bibliographystyle{ieeetr} 


\end{document}